\documentclass[preprint,12pt,authoryear]{elsarticle}

\makeatletter
\def\ps@pprintTitle{%
  \let\@oddhead\@empty
  \let\@evenhead\@empty
  \let\@oddfoot\@empty
  \let\@evenfoot\@oddfoot
}
\makeatother

\usepackage{amssymb}
\usepackage{color}
\usepackage{xcolor}
\usepackage{amsmath}
\usepackage{graphicx, wrapfig}
\usepackage{soul}
\usepackage{subfig}
\usepackage{hyperref}
\usepackage{etoolbox}

\journal{Neural Networks}

\begin{document}
\begin{frontmatter}

\title{Event Fusion Photometric Stereo Network}

\author[inst1]{Wonjeong Ryoo}
\author[inst3]{Giljoo Nam}
\author[inst2]{Jae-Sang Hyun\corref{mycorrespondingauthor}}
\cortext[mycorrespondingauthor]{Corresponding authors}
\author[inst1]{Sangpil Kim\corref{mycorrespondingauthor}}
\address[inst1]{Department of Artificial Intelligence, Korea University, Seoul, 02841, South Korea}
\address[inst2]{School of Mechanical Engineering, Yonsei University, Seoul, 03722, South Korea}
\address[inst3]{Reality Labs Research, Meta, USA}

\begin{abstract}
We present a novel method to estimate the surface normal of an object in an ambient light environment using RGB and event cameras.
Modern photometric stereo methods rely on an RGB camera, mainly in a dark room, to avoid ambient illumination.
To alleviate the limitations of the darkroom environment and to use essential light information, we employ an event camera with a high dynamic range and low latency. 
This is the first study that uses an event camera for the photometric stereo task, which works on continuous light sources and ambient light environment. 
In this work, we also curate a novel photometric stereo dataset that is constructed by capturing objects with event and RGB cameras under numerous ambient lights environment.
Additionally, we propose a novel framework named Event Fusion Photometric Stereo Network~(EFPS-Net), which estimates the surface normals of an object using both RGB frames and event signals.
Our proposed method interpolates event observation maps that generate light information with sparse event signals to acquire fluent light information.
Subsequently, the event-interpolated observation maps are fused with the RGB observation maps. 
Our numerous experiments showed that EFPS-Net outperforms state-of-the-art methods on a dataset captured in the real world where ambient lights exist.
Consequently, we demonstrate that incorporating additional modalities with EFPS-Net alleviates the limitations that occurred from ambient illumination.
\end{abstract}






\end{frontmatter}

\section{Introduction}
Photometric stereo~\citep{woodham1980photometric} is an important computer vision task that estimates the surface normals of a real object by observing the reflections of light sources. 
The ideal photometric stereo method adopts a darkroom setting to avoid ambient illumination and assumes a Lambertian reflection of the object's surface. 
However, because Lambertian surface objects are impossible to obtain in real-world scenarios, several works~\citep{ikehata2012robust,ikehata2014photometric,santo2017deep,ikehata2018cnn,chen2018ps,logothetis2021px,zheng2019spline} have exploited non-Lambertain objects to approximate the real world. 

To fully model the real-world environment, recent works~\citep{hung2015photometric, hold2019single} have additionally considered  ambient light environments for surface normal estimation. Unlike the studies that utilized only the main light source, these studies used all light sources illuminated on the object. It is difficult to compute the light source information in an ambient light environment because ambient illumination on the surface causes saturation, which is over the pixel intensity value range of expression on the frame. Moreover, these methods fail to precisely predict the surface normal in the wild because they cannot accurately extract essential light information and use approximation methods or contain an error range.

Generally, photometric stereo uses RGB cameras. However, RGB cameras have a limited dynamic range. A limited dynamic range causes saturation of pixel intensities, and hence RGB cameras cannot represent light source information completely. Additionally, it is difficult to exclusively capture the effect of the main light source because all the pixels in the frame update their intensities each time a camera is triggered. In this study, we aimed to overcome these limitations of RGB cameras and improve the surface normal estimation in an ambient light environment by incorporating an event camera, similar to \citep{lichtsteiner2008128, posch2010qvga, brandli2014240}. The notable advantages of the event camera are its sensitivity to small changes in light intensity, high dynamic range, and high temporal resolution. The event signals are captured asynchronously when the change in light intensity exceeds a certain threshold. Accordingly, an event camera is useful for capturing sensitive changes in light intensities due to the movement of the main light in the photometric stereo task. Unlike an RGB camera, an event camera can acquire the necessary light source information more accurately and precisely, even if only the main light direction is used in the ambient light setting.

Event signals occur from the relative values of the light intensities. Therefore, we need absolute light intensities, which can be the standard, such as RGB frames. So, we incorporated RGB frames and event camera signals. The event camera receives information asynchronously, whereas the RGB camera captures the scene synchronously. More specifically, the event signal consists of pixel coordinates, time stamps, and polarity, which express the events in the scene. Thus, asynchronous event signals cannot be fused directly with RGB camera frames without appropriate fusion methods. To adequately combine the advantages of each camera, there are additional preprocessing approaches~\citep{rebecq2017real,innocenti2021temporal,gehrig2019end} to match the event signals to a voxel grid. A voxel grid consisting of accumulated polarities in each pixel of several channels divided into uniform cycles was used to construct the frame-like arrays. Inspired by existing approaches~\citep{rebecq2017real,innocenti2021temporal,gehrig2019end} to represent event signals with a voxel grid, we introduce a fusion method to fuse the two heterogeneous camera signal features to deal with heterogeneous modalities. Our fusion method represents light source information in a complex manner.

Consequently, we introduce an Event Fusion Photometric Stereo Network~(EFPS-Net) to utilize RGB frames and event signals. Unlike dense RGB frames, event signals that only occur when the light intensity change is over the threshold produce sparse data. To reduce the density gap between event signals and RGB frame, we propose a novel interpolation network that interpolates the sparse observation maps~\citep{ikehata2018cnn} generated with event signals to make them similar to the dense observation map generated by RGB frames. Subsequently, the observation fusion module fuses four observation maps generated with the RGB camera and one interpolated observation map generated by the event camera to complement the light source information of each observation map. The features consisting of observation maps generated from the two cameras achieve better performance in predicting the surface normal than using only one modality feature by complementing each disadvantage.

To show event signals that contain information on light sources, we curate an RGB-event paired dataset, that is recorded in an ambient light environment with an RGB camera and an event camera. Our proposed method outperforms other state-of-the-art methods that use an RGB camera on real datasets. To the best of our knowledge, our proposed method is the first work to fuse RGB and event signals for photometric stereo tasks. From our extensive experimental results, utilizing event signals along with RGB frames is much more effective in photometric stereo tasks under ambient light conditions than using only an RGB camera.

In this paper, our contributions are summarized as follows:
\begin{itemize}
    \item We introduce a novel method called EFPS-Net that can solve the real-world photometric stereo problem in the wild environment with existing ambient lights. 
    \item We propose a technique to interpolate the observation map from event signals that represent dense light information than prior methods.
    \item Our methods perform fusion between the RGB-based observation maps and the event interpolated observation map to predict the surface normal, especially the edge.
    \item We curate the RGB event pair dataset made using an RGB camera and an event camera in a real-world environment where ambient light exists, and we show that our method is the state-of-the-art method when using event signals.
\end{itemize}

\section{Related works}
\subsection{Photometric stereo}
Photometric stereo estimates the surface normals of the local geometry of an object using at least three light sources and frames from each light source. In particular, the object must be a Lambertian reflectance surface to use a diffuse Bidirectional Reflectance Distribution Function~(BRDF). 
Perfectly diffuse BRDF cannot exist in the real world because real-world environments violate these constraints. Especially real-world objects have non-Lambertian reflectance surfaces, which are affected by global illumination effects such as ambient lights and casting shadows. When using only an RGB camera, we cannot know the effects of the main light on an object in an ambient light environment. Thus, existing methods~\citep{hung2015photometric, hold2019single} convert all the light information that affects an object into a feature, such as one light information or learnable parameter. However, it contains unnecessary ambient light information. Moreover, the pixel intensities and reflectance directions in each light direction are quite different depending on the components of the object and sure
face properties. Therefore, there are various BRDFs based on reflectance. Since the introduction of photometric stereo by Woodam~\citep{woodham1980photometric}, various methods have been proposed that are robust to these external constraints. Notably, the former methods are the two main categories of traditional and learning-based methods.

First, traditional methods attempt to solve the non-Lambertian reflectance problem statistically~\citep{ikehata2012robust,ikehata2014photometric} or use a sophisticated reflectance model \citep{oren1995generalization,hertzmann2003shape,alldrin2008photometric,goldman2009shape}. A statistical method~\citep{ikehata2012robust} adopts a hierarchical Bayesian approximation. This method achieves competitive performance in estimating surface normals even with a non-Lambertian reflectance surface. However, such methods require many frames and handling high-frequency materials poses substantial difficulties. To overcome these requirements, another method~\citep{ikehata2014photometric} was proposed that adopts a piece-wise linear diffuse model to get better performance. Despite these efforts to improve the non-Lambertian reflectance problem, these still fail to overcome global illumination effects in large parts of objects. Additionally, these methods require more improvement to estimate surface normals precisely when several light sources are available.

To resolve the constraints not solved by traditional methods, learning-based methods using neural networks have been researched. The Deep Photometric Stereo Network~(DPSN)~\citep{santo2017deep}, which is the first neural network method, has achieved better performance than other traditional methods in real-world photometric stereo. However, the DPSN must be fixed and it must have the same light directions during the training and testing phase. To solve the DPSN problem, PS-FCN~\citep{chen2018ps} trains and tests with random order input, which consists of a pair of RGB frames and light directions reshaped in the RGB frame representation. However, PS-FCN needs a lot of train data and cannot fuse intensities effectively when the frames have many saturated signals because it uses a max pooling layer to fuse intensities. Another method is CNN-PS~\citep{ikehata2018cnn}, which estimates surface normals utilizing an observation map that represents light information with intensities per light direction captured from each pixel. Unlike CNN-PS, which generates an observation map from grayscale frames, another method~\citep{logothetis2021px} uses fluent light information by generating observation maps from each channel of the frames and averaging them. These approaches fail to predict the surface normal of an area in the object under limited light conditions or when ambient illumination environments. Consequently, these problems cannot be resolved using only an RGB camera because of the lack of light information and interruption of ambient illumination. To fundamentally overcome this problem, we used an event camera that can capture a more diverse range of wider dynamic range of lights than that captured by the RGB camera.

\subsection{Event camera}
The mechanism of an event camera is inspired by the human optic nerve system, resulting in low power consumption, high dynamic range, and significantly low latency~\citep{gallego2020event}. A conventional RGB camera captures light signals using a two-dimensional~(2D) plan sensor and saves them as an array.

However, in an event camera, event signals are captured pixel-by-pixel asynchronously, along with the time stamp and polarity representing the on and off terms.
Moreover, the event camera disregards the ambient illumination. It primarily detects the change in intensity of the main light because it captures the light intensity differences of each pixel on a logarithmic scale. However, event signals are not suitable for neural networks without proper post-processing. Event signals are not detected when there are no changes in scene lighting, making the data sparse when the changes are minimal.
The format of the event signal is similar to that of point clouds, which are a list of multiple data points consisting of three components: pixel coordinates, time stamp, and polarity.
To use event signals effectively with neural networks, numerous representation methods have been proposed utilizing event signals such as graphs~\citep{bi2019graph,bi2020graph}, point clouds~\citep{sekikawa2019eventnet,vemprala2021representation}, spikes~\citep{kim2021optimizing,lee2020spike}, frames~\citep{rebecq2017real,innocenti2021temporal,gehrig2019end,ryan2021real}, patches~\citep{sabater2022event}, or etc~\citep{dong2022event}. 

Several approaches~\citep{messikommer2022multi,gehrig2021combining,messikommer2022bridging} incorporate the two modalities~(i.e., RGB and event cameras) and use event representation methods in various tasks. Consequently, considerable advantages are achieved by compensating for the disadvantages of each modality. The RGB camera expresses absolute light intensities but operates insufficiently in saturated or slightly exposed parts. Event cameras represent relative light intensities, even saturated or underexposed parts, in RGB cameras. Despite providing different outputs, the two modalities compensate for each other since the RGB frames and event signals information overlap substantially because the two cameras capture the same scene irradiance. Thus, the output of each camera can account for the information missing in the other's output. Their combined use can satisfactorily overcome ambient illumination environments in photometric stereo.

\section{Methods}
\begin{figure}[t]
    \centering
    \includegraphics[width=\textwidth]{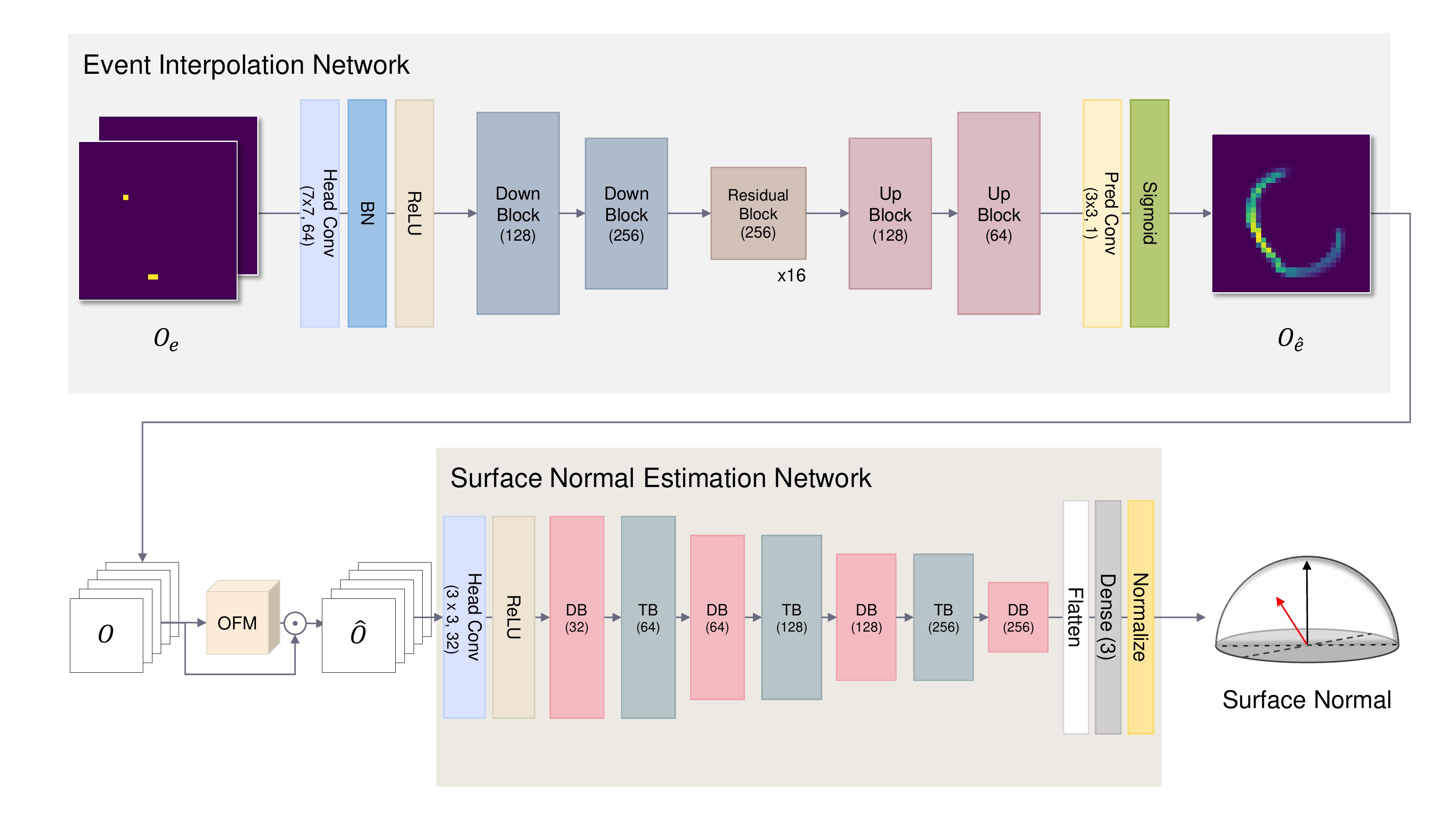}
    \caption{Overview of Event Fusion Photometric Stereo Network~(EFPS-Net). EFPS-Net consists of three parts: (i)~Event Interpolation Network~(EI-Net) which interpolates the sparse event observation maps~$O_{e}$,~(ii) Observation Fusion Module~(OFM) that fuses five observation maps, and~(iii) Surface Normal Estimation Network~(SNE-Net) which outputs surface normal vector given fused observation maps.}
    \label{fig:modelarchitecture}
\end{figure}
In this section, we describe our proposed pipeline, i.e., Event Fusion Photometric Stereo Network~(EFPS-Net) shown in Fig.~\ref{fig:modelarchitecture}, which utilizes fused information from RGB frames and event signals.

\subsection{Obseravation map}
An observation map is a feature map that consists of light intensities from each direction.
More specifically, To compute the observation map~$O_{c}$ of ${m\times m}$ size for channel~$c$, we project the $j$~th intensity~$i_{j_{c}}$ of light direction~$(l^{x}_{j},~l^{y}_{j},~l^{z}_{j})$ from the shaped unit hemisphere onto the 2D plane, as defined in Eq.~\ref{input:observationmap}.

\begin{equation}
O_{c}\left( \left \lfloor m\frac{l^{x}_{j}+1}{2}\right \rfloor,\left \lfloor m\frac{l^{y}_{j}+1}{2}\right \rfloor \right) = i_{j_{c}}
\label{input:observationmap}
\end{equation}
Since, at a time, this approach only deals with the necessary part of pixels in either RGB frames or event signals, it is computationally efficient. To fuse two different modalities effectively with observation maps, we converted RGB frames into observation maps~$O_{r,g,b,n}$ and transformed raw event signals into sparse event observation maps.  
In particular, we utilize the method from PX-Net~\citep{logothetis2021px}, which generates RGB observation maps~$O_{r,g,b}$ from RGB frames and $O_{n}$ which is a normalized map of RGB observation maps as defined in Eq.~\ref{input:normalize} to obtain $O_{r,g,b,n}$.

\begin{equation}
O_{n} = \frac{O_{r} + O_{g} + O_{b}}{\text{max}(O_{r} + O_{g} + O_{b})}
\label{input:normalize}
\end{equation}

\subsection{Event Interpolation Network}
To extract useful information from the event signals for EFPS-Net, we propose a novel event signal representation that can also be transformed into an observation map.
The proposed method is fundamentally different from previous representation methods, which simply accumulate the polarity along the time at each pixel coordinate.
We divided the polarities by $\lambda$ before accumulating generated event signals to ensure a wide distribution and applied a non-linear function such as a hyperbolic tangent.
Subsequently, the processed event signals were located in the voxel grid.
Moreover, we separated polarity into two channels.
One channel represents a positive polarity, which indicates that the light intensity is increased. 
The other channel represents negative polarity, which indicates decreased light intensity.
Consequently, we propose a new separate event voxel grid~$V_{s}$ containing the dimensions of polarity, as defined in Eq.~{\ref{input:separate_voxel_grid}}.
We generated $O_{e}$ with each polarity channel of $V_{s}$ after merging all the time slots.

We note that the event observation maps are sparse because event signals do not compulsorily occur even if the light direction is changed.
Sparse event observation maps are insufficient for estimating the surface normal.
Therefore, we used an EI-Net to interpolate the event observation maps into a dense normalized observation map that contains sufficient information.

$$
V\rightarrow \begin{cases}
V(x,y,\delta,0) = \sum^{k(\delta+1)}_{t_{n}=k\delta}{\frac{1}{\lambda}}, & \text{ if } p(x,y,t_{n})=1. \\
V(x,y,\delta,1) = \sum^{k(\delta+1)}_{t_{n}=k\delta}{\frac{1}{\lambda}}, & \text{ if } p(x,y,t_{n})=-1.
\end{cases}
$$
where $B$ is the channel of voxel grids generated during $\Delta T$, which is one period. One channel of voxel grid~$k$ indicates $\frac{\Delta T}{B}$.

\begin{equation}
V_{s} = \frac{\text{exp}^{V}-\text{exp}^{-V}}{\text{exp}^{V}+\text{exp}^{-V}}
\label{input:separate_voxel_grid}
\end{equation}

\begin{figure}[t]
    \centering
    \subfloat[]{
    \includegraphics[width=0.475\textwidth]{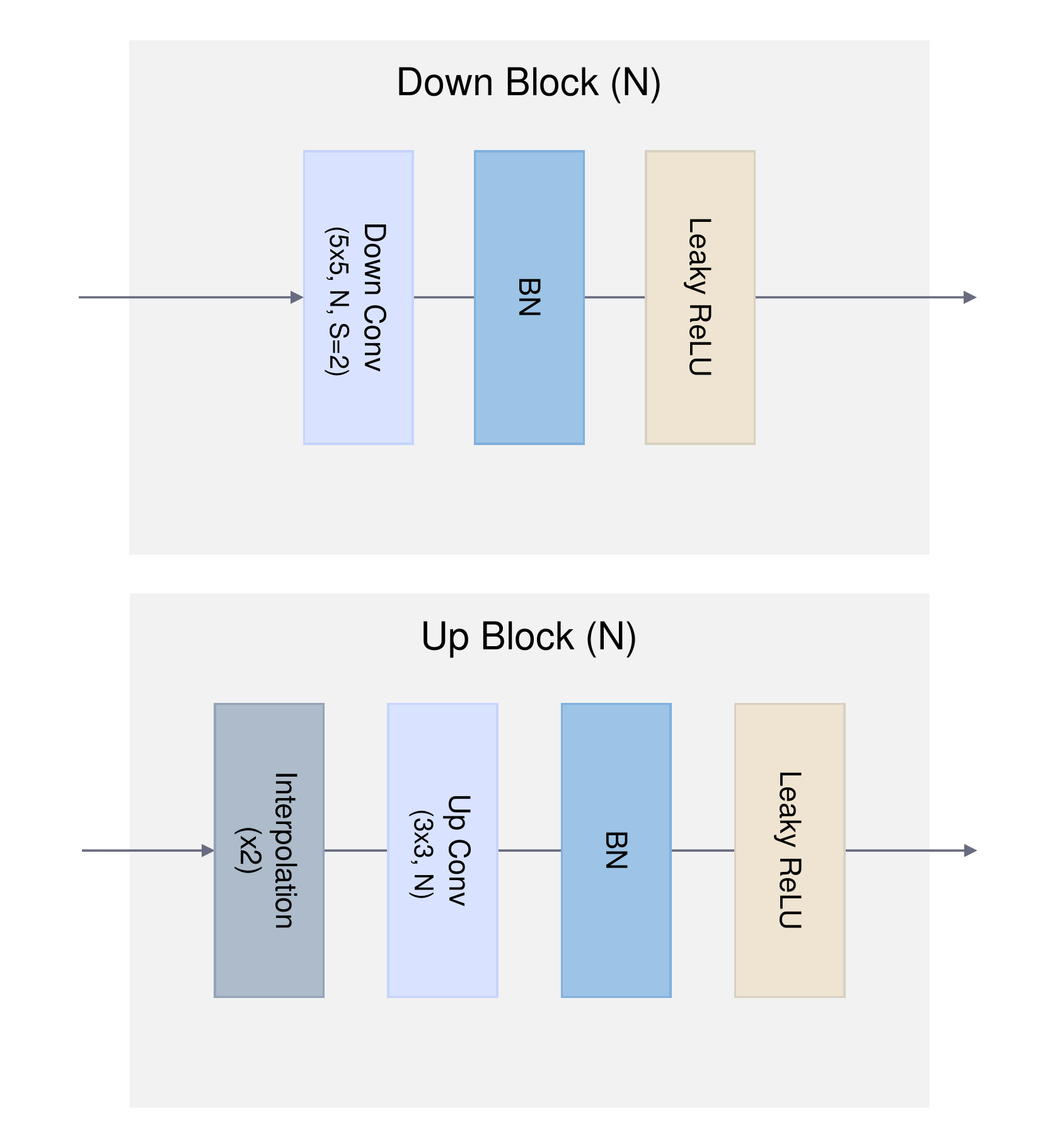}
    }
    \subfloat[]{
    \includegraphics[width=0.475\textwidth]{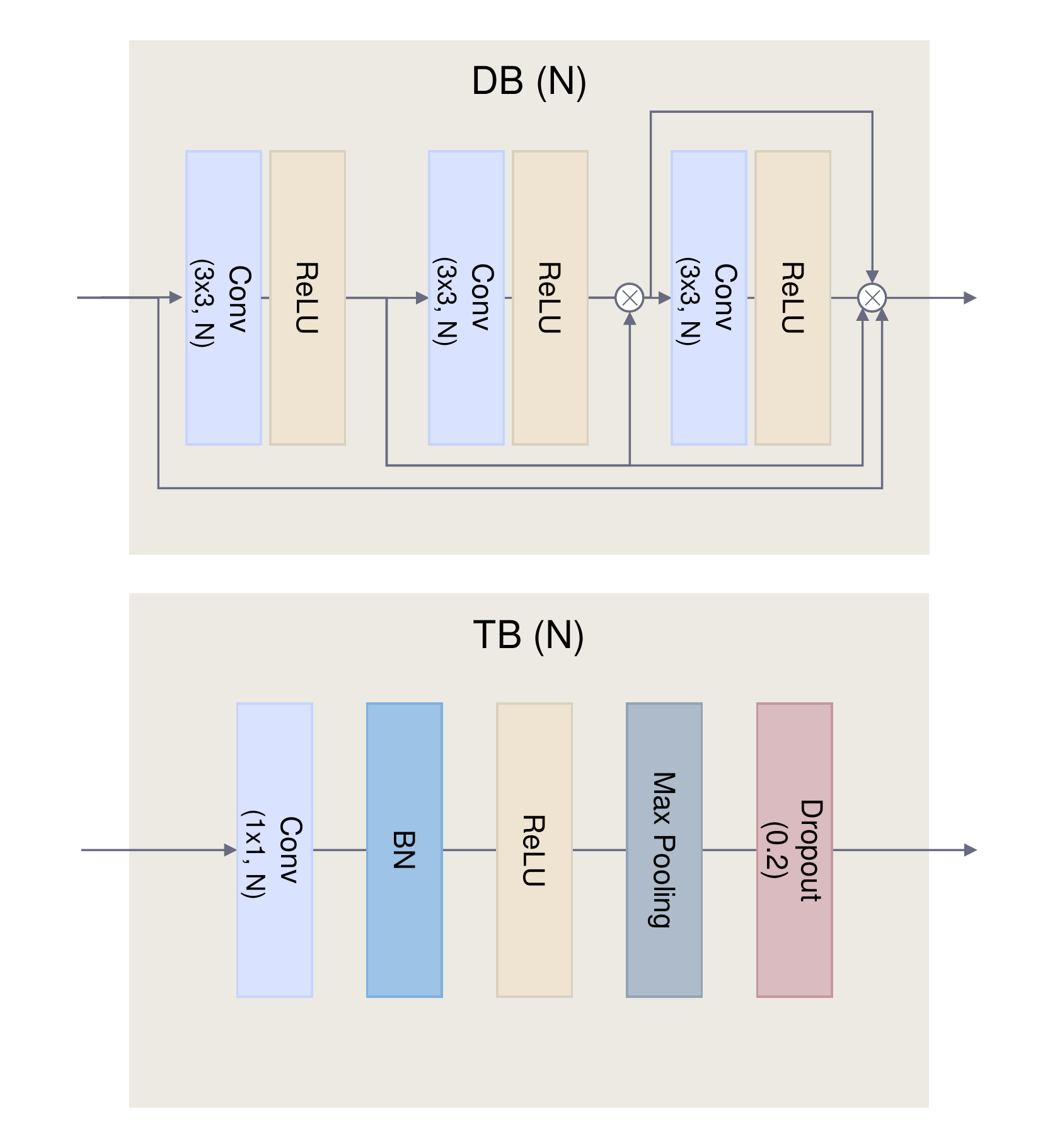}
    }
    \caption{Details of Event Interpolation Network~(EI-Net) blocks and Surface Normal Estimation Network~(SNE-Net) blocks. First, EI-Net blocks, depicted in (a), have two blocks which are Down Block encoding features and Up Block up-sampling observation map. Next, SNE-Net blocks consist of DB~(Dense Block) and TB~(Transition Block) as shown in (b). DB is same as \citep{logothetis2021px}, but TB contains an additional batch normalization layer.}
    \label{fig:blocks}
\end{figure}

EI-Net consists of a head part, two Down blocks, sixteen Residual blocks, two Up blocks, and follow by a pred part.
The head part is composed of a head convolution layer, batch normalization layer, and ReLU activation function.
This part amplifies the sparse event observation map to multiple channels, and the Down Block encodes them while reducing the size by half.
As shown in Fig.~\ref{fig:blocks}~(a), the Down Block comprises $5\times5$ kernel convolution layer, a batch normalization layer, and a leaky ReLU activation function.
After obtaining encoded feature maps from the last Down Block, sixteen residual blocks deal with the output feature maps to interpolate the sparse intensity features.
As expressed in Eq.~\ref{input:residual_block}, the residual block consists of a $3\times3$ kernel convolution layer, a batch normalization layer, and a ReLU activation function.
It derives more fluent-intensity output feature maps~$M_{out}$ from the sparse input feature maps~$M_{in}$.
To decode resized inputs, we used Up Blocks, shown in Fig.~\ref{fig:blocks}~(a), which upscale the last layer output of the residual block's $M_{out}$ to match the EI-Net's input resolution of EI-Net.
Finally, we transformed them into an interpolated event observation map using the pred part, which consists of a pred convolution layer and adjusts values between 0 and 1 using a sigmoid activation function to match the range of input values.

\begin{equation}
M_\text{out} = \text{ReLU}(\text{BN}(f^{3 \times 3}(M_\text{in}))) + M_\text{in}
\label{input:residual_block}
\end{equation}

\subsection{Observation Fusion Module}
\begin{figure}[t]
    \centering
    \includegraphics[width=0.5\textwidth]{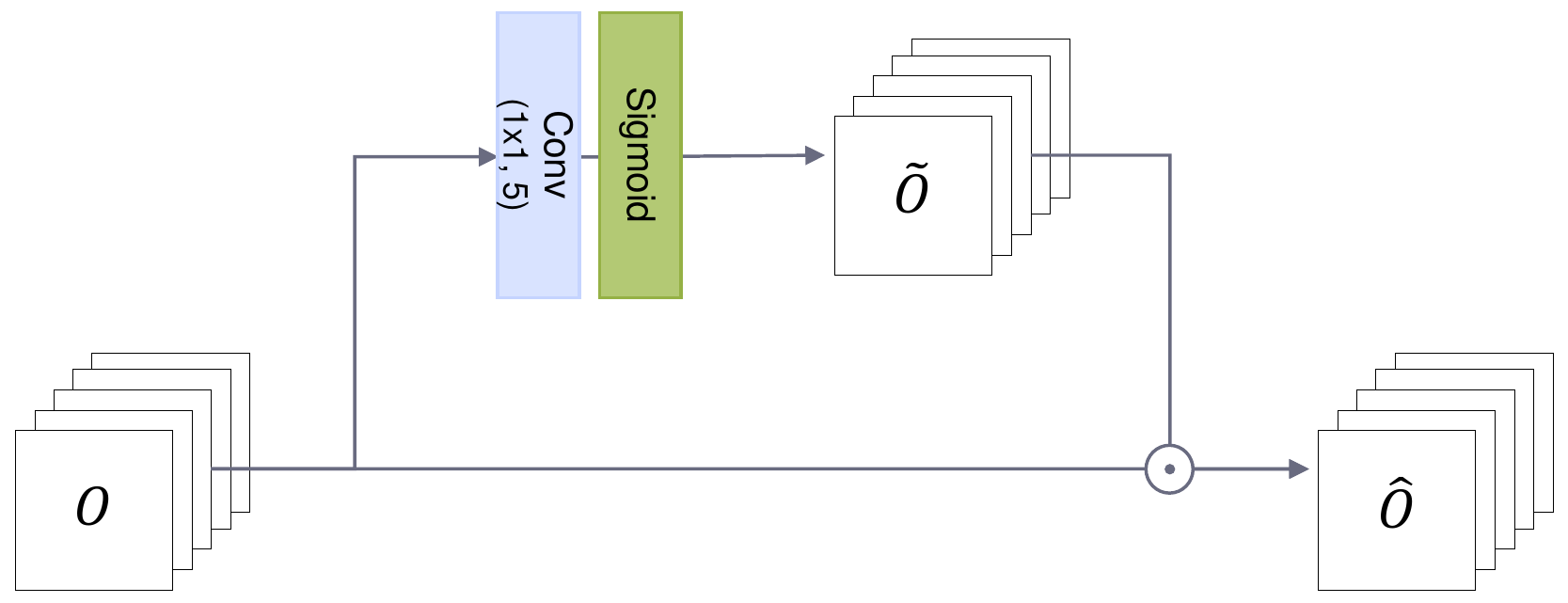}
    \caption{Overview of Observation Fusion Module~(OFM), which produces five refined observation maps using convolution layers.}
    \label{fig:ofm}
\end{figure}

In our model architecture, the SNE-Net utilizes fused observation maps~$\hat{O}$ as the input.
To acquire fused observation maps, we gathered five new observation maps~$\tilde{O}$ with point-wise convolution.
This operation compensates for each observation map by using other observation maps.
Subsequently, we performed element-wise multiplication for each channel on the OFM input~$O$ after setting it in the 0 to 1 range with the sigmoid activation function.
It injects interacted observation map features at the input observation maps that were concatenated with $O_{r,g,b,n}$ and $O_{\hat{e}}$.
Consequently, we acquire $\hat{O}$, complemented by each channel of the observation map, as input to SNE-Net.
Overall, the OFM implements the process illustrated in Fig.~\ref{fig:ofm}.

\subsection{Surface Normal Estimation Network}
SNE-Net is inspired by PX-Net~\citep{logothetis2021px}, which uses an observation map. SNE-Net uses a large batch size because the observation map size is small. When the model has a large batch size, appropriate batch normalization layers are especially effective. Thus, we attached the batch normalization layer~(BN) to our network between the convolution layer and the ReLU activation function in the Transition Block~(TB), as shown in Fig.~\ref{fig:blocks}~(b). 
Finally, SNE-Net estimates the surface normal from fused observation maps generated from the OFM.

\subsection{Loss function}
EFPS-Net estimates the surface normal per pixel on an object using numerous light directions and intensities represented by image frames and event signals according to each light direction. We used the scale-invariant loss~$\mathcal{L}_{e}$ Eq.~(\ref{input:scale-invariant})~\citep{eigen2014depth} to interpolate the sparse event observation maps~$O_{e}$ to the dense normalized observation map~$O_{n}$. Since event signals represent relative light information based on the absolute intensity of RGB frames, we interpolated them to $O_{n}$.

\begin{equation}
\mathcal{L}_{e} = \frac{1}{n}\sum_{i}R^{2}_{i}-\frac{1}{n^{2}}(\sum_{i}R_{i})^{2},
\label{input:scale-invariant}
\end{equation}
where $R$ is computed using $R_{i} = O_{\hat{e}(i)} - O_{n(i)}$.\\

Moreover, we applied the Mean Angular Error~(MAE) loss function defined in Eq.~\ref{input:mae}. This loss function optimizes the error between the ground truth surface normal~$\textbf{n}$ and the predicted surface normal~$\hat{\textbf{n}}$.

\begin{equation}
\mathcal{L}_{n} = \cos^{-1}(\textbf{n}\cdot \hat{\textbf{n}})
\label{input:mae}
\end{equation}

In summary, EFPS-Net is optimized by combining loss~$\mathcal{L}_{e}$ and loss~$\mathcal{L}_{n}$, defined as:
\begin{equation}
\mathcal{L}_{total} = \mathcal{L}_{e} + \mathcal{L}_{n}
\label{input:total}
\end{equation}

\section{Dataset curation}
\subsection{Data acquisition system}
Typically, photometric stereo datasets~\citep{chen2018ps,ikehata2018cnn,shi2016benchmark} are collected using RGB frames owing to their availability.
Despite the convenience of image collection using an RGB camera, the datasets have some issues related to saturation and a limited dynamic range. To overcome these limitations, we introduce a novel data acquisition system with an event camera for photometric stereo datasets in an ambient light environment. The configuration of RGB and event cameras mitigates the saturation problem by extending the dynamic range.

\begin{figure}[t]
    \centering
    \includegraphics[width=0.475\textwidth]{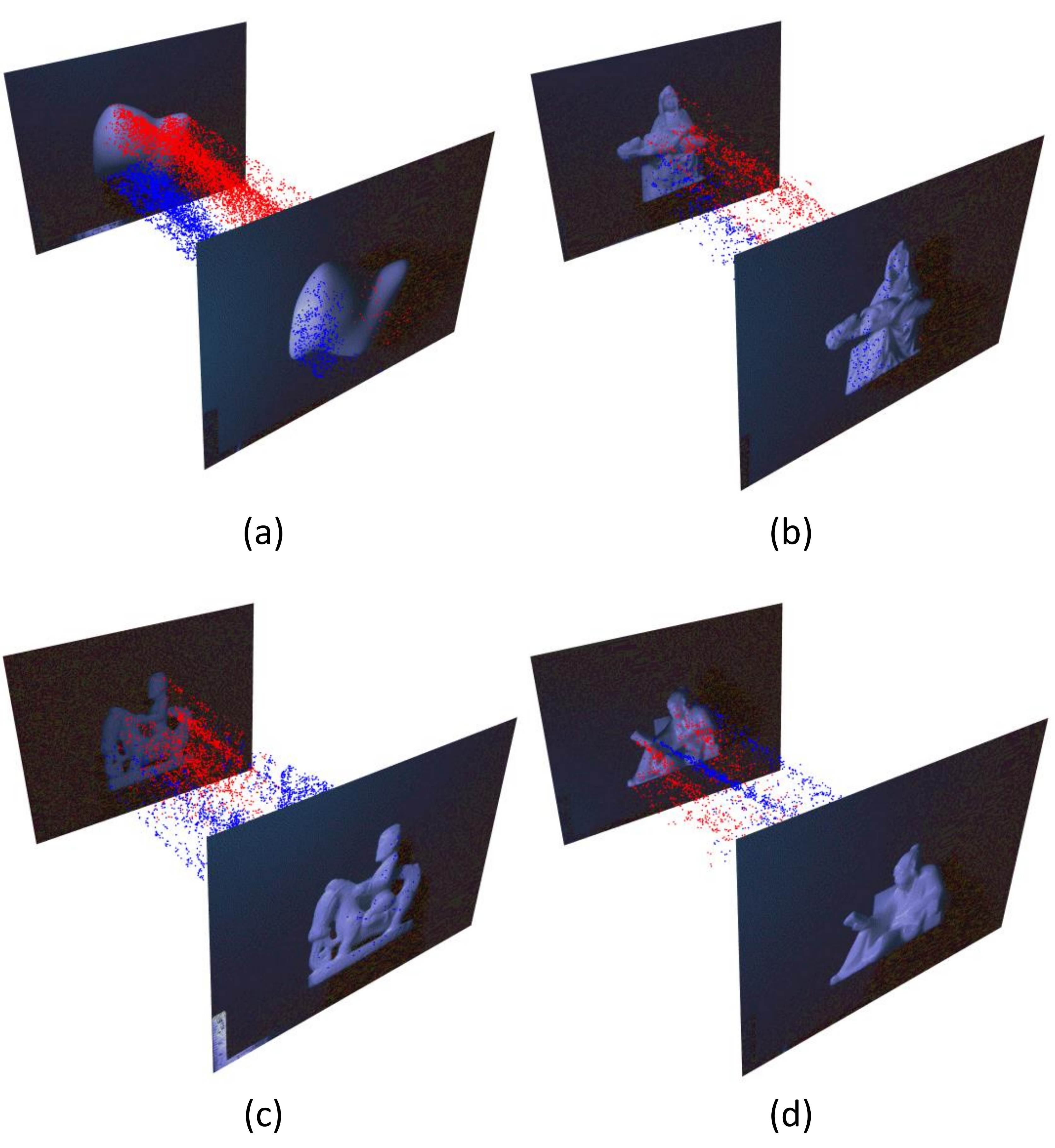}
    \caption{Examples of event signals. (a) is blob09, (b) is sculpture04, (c) is horseman, and (d) is reading. (a), (b), (c) are parts of the training dataset, and (d) is part of the test dataset.}
    \label{fig:event_flow}
\end{figure}
There are two methods for generating event data: synthetic data creation using grayscale images~\citep{hu2021v2e} and real-world data creation using an event camera. The synthetic data creation method generates event signals based on the concept of how many pixels change in two frame intervals. Even if the change in each pixel is synthesized well, the majority of the event signals are generated around the starting time stamp. Thus, the representation of the change in pixel values decreases as it reaches the end of each time stamp. To avoid this uneven distribution of synthetic data, we used an event camera to fill the gap within the time intervals. Fig.~\ref{fig:event_flow} shows the event signals during a specific period in the two frames. To use the temporal property of event signals, our dataset was generated with lights that moved continuously and, were not fixed. The exact light trajectory cannot be reproduced over time, and the light intensity affecting the object changed every moment. Therefore, we used RGB and event cameras simultaneously. There are two main approaches for gathering datasets. The first involves recording the 3D model with two cameras to obtain RGB frames and event signals and acquire ground-truth normals with a 3D scanner. Another method is to collect a 3D model printed with a 3D printer and acquire ground-truth normals after matching the camera settings in Blender~\citep{blender} with the experimental camera settings. We adopted the latter method to obtain ground-truth normals with bare effort.

Conventional photometric stereo methods collect datasets for all main lights by fixing them, turning on the main lights individually, and then recording them. In contrast, event signals have a continuous property that represents uninterrupted light changes from nearby pixels. Since the main light source trajectories change as the light moves, a high-speed RGB camera is required to capture light movements. We used a Grasshopper 3 USB3~(GS3-U3-51S5C-C) RGB camera operating with a global shutter that exposes all pixels to the light simultaneously and a DAVIS 346 mono event camera. The maximum resolution is $2448 \times 2048$, and the maximum frame rate with this resolution setting is 75~FPS. To increase the camera capturing speed, we slightly reduced the resolution to $1760 \times 1600$ at 90~FPS. By contrast, the event camera resolution is $346 \times 260$, generating event signals and 30~FPS grayscale frames. We turned off all functions that adjust the exposure, white balance, and gain values automatically and manually set them with specific values. In addition, we used KOWA~(LM12JC1MS) lens, of which the focal length is 12~mm for the RGB camera and 16~mm focal lengths KOWA~(LM16JC1MS) lens for the event camera. To ensure both cameras capture the object(s) as closely as possible, we used lenses with different focal lengths and field of view~(FOV). When the image plane is transformed with the settings, the occluded area is reduced significantly. Subsequently, we minimized the effect of the radial distortion created by the non-linearity of the lenses by applying the coefficients of the lens distortion model in \ref{sec:equations}.

\begin{figure}[t]
    \centering
    \includegraphics[width=0.475\textwidth]{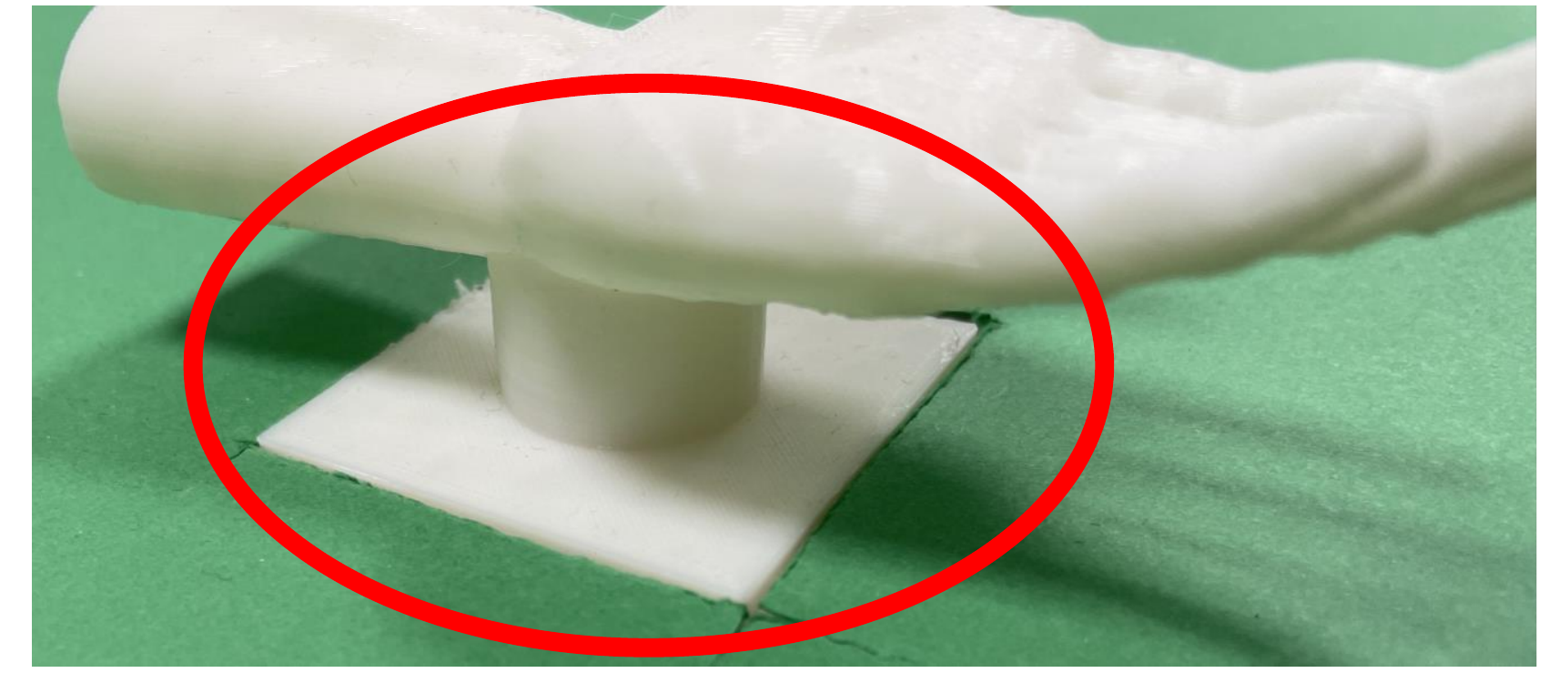}
    \caption{Red circle indicates the pedestal that upholds and fixes the object.}
    \label{fig:Pedestal}
\end{figure}
\begin{figure}[t]
    \centering
    \includegraphics[width=0.95\textwidth]{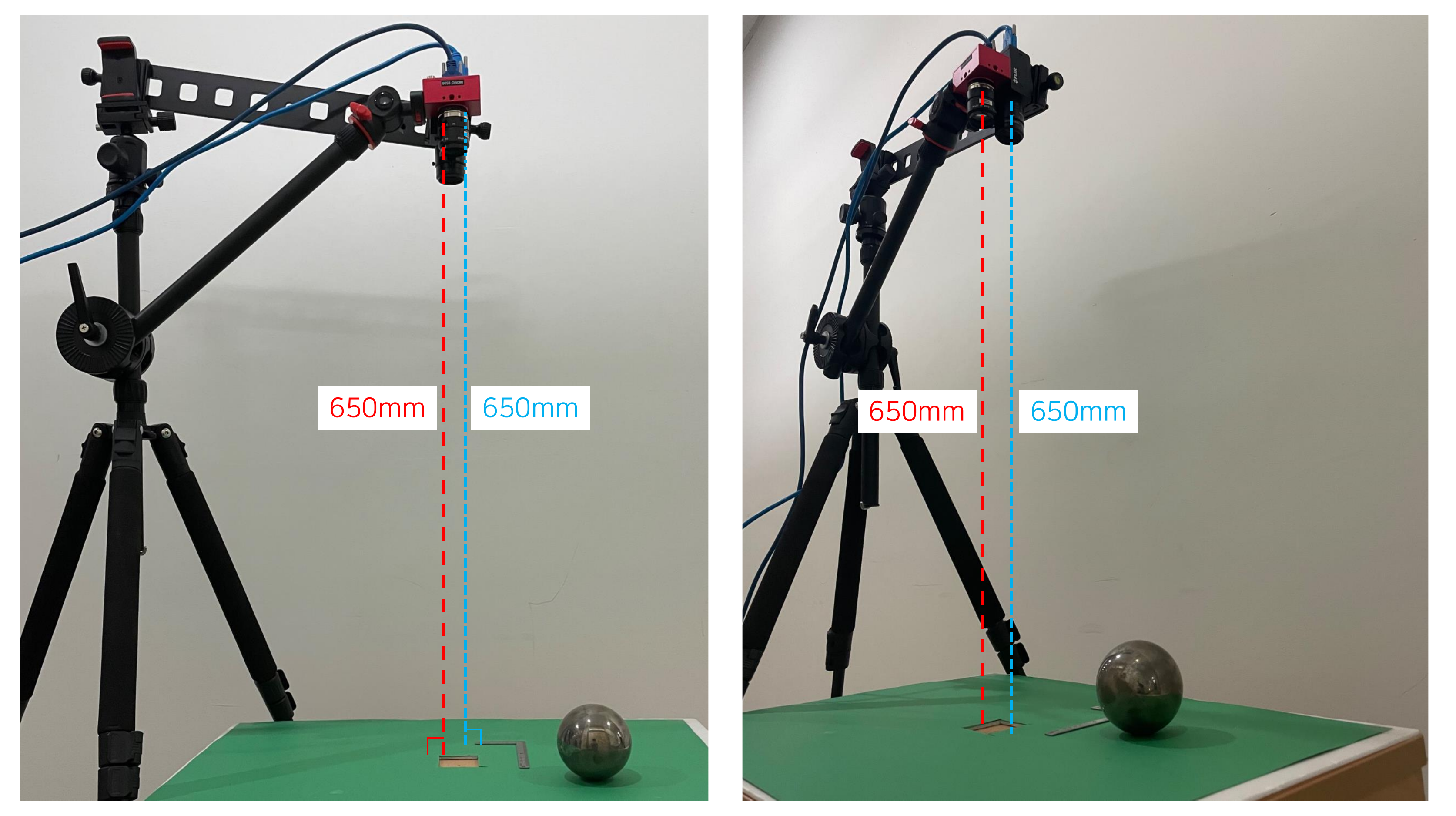}
    \caption{Our experiment setting. The set consists of a tripod with the fixed event camera and the RGB camera, an experimental table, a square, and a chrome ball. The red line is the distance between the event camera and the experimental table, and the blue line is the distance between the RGB camera and the experimental table.}
    \label{fig:setting}
\end{figure}

To obtain the data, we printed 23 pieces of 3D models for the training dataset and 10 pieces of 3D models for the test dataset with a 3D printer. The 3D model was rescaled to make all models normalized to the size of 100~mm among width, length, and height. It is challenging to represent a 3D model with a synthetic world coordinate system because there is no angle information with respect to the world coordinates. Therefore, to fix the 3D model, we printed a square pedestal, cylindrical column, and 3D model simultaneously, as indicated by the red circle in Fig.~\ref{fig:Pedestal}. A pedestal makes it easier to map the 3D model in the real world to a synthetic world coordinate system. Subsequently, we drilled a hole with a square pedestal in a table and fixed the 3D model to the hole. Both cameras were set 650~mm above the experimental table to obtain data from the center point of the hole. Additionally, to obtain the light directions for each 3D model, we used a chrome ball with a radius of 35~mm near the 3D model. Fig.~\ref{fig:setting} shows the experimental setting.

We created a dataset for an indoor environment containing several ceiling lamps. Moreover, we used both RGB and event cameras to record 150 frames of each 3D model as light trajectories, created by hand, moved around the object. The RGB frames and event signals were incorporated into the RGB event pair dataset for one 3D model. To obtain the ground-truth surface normals of this dataset, the 3D model captured by the event camera should be matched with the 3D model from the synthetic environment created by Blender. To match the area of the captured 3D model, we set the elements of the camera, lens, and distance between the camera and the 3D model in Blender, similar to the experimental setting. Additionally, we set the scale, location, and rotation of the 3D model to be the same as those of the 3D model in the captured image. The 3D model was fixed with a pedestal in the real world, mapping easily to a synthetic world coordinate system by quickly determining the scale, location, and rotation ideal values of the 3D model to use in world coordinates. A setting similar to the real world was first set up in Blender, and then we acquired the ground-truth surface normals and segmentation masks with Vision Blender~\citep{cartucho2020visionblender}, which is an add-on in Blender.

\begin{figure}[t]
    \centering
    \includegraphics[width=0.95\textwidth]{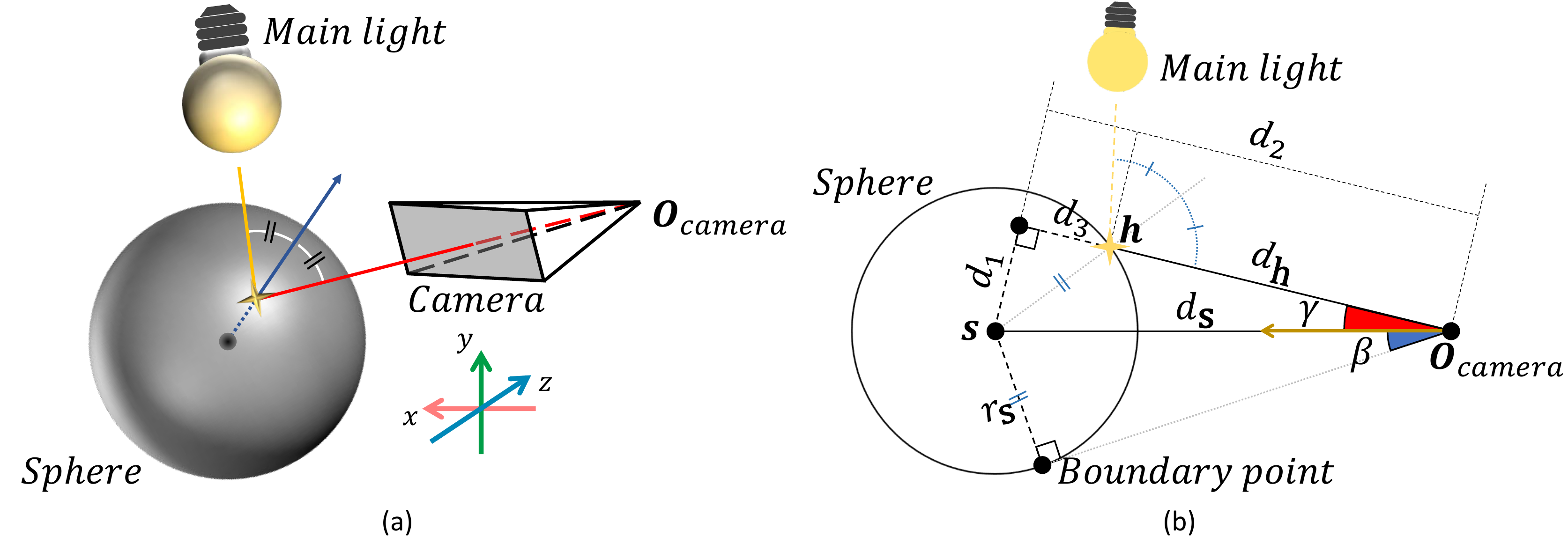}
    \caption{Determination of light direction. Approach to get light direction indicates on 3 Dimensional space in (a) and 2 Dimensional plane in (b).}
    \label{fig:lightdirection}
\end{figure}
Finally, to obtain the light directions, we tracked the main light using a chrome ball recorded from the RGB camera. To explain this process briefly, we use the bisector property. In our case, the direction~$\textbf{N}$ from the center of the chrome ball to the highlight point on the chrome ball is the bisector. It separates the center between light direction~$\textbf{L}$ and viewing direction~$\textbf{R}$. This process is illustrated in Fig.~\ref{fig:lightdirection} (a).

Specifically, we masked all areas of the frame captured by the RGB camera, except for the chrome ball part. The main light recorded on the chrome ball was saturated. Therefore, we masked the saturated parts again, except for the main light. Subsequently, we expressed the main light with a point that averaged the values of the saturated coordinates in a masked RGB frame. The point about the main light coordinate is transformed into a camera-based world coordinate system using the camera parameters. The main light coordinate, as well as the coordinates of the chrome ball diameter endpoints, and chrome ball center were transformed into a camera-based world coordinate system. Assuming the chrome ball radius is $r_{\textbf{s}}$, we indicate each value in Fig.~\ref{fig:lightdirection} (b) using Eq.~\ref{input:figureld}.
\begin{equation}
\begin{aligned}
d_{\textbf{s}} &= r_{\textbf{s}}/\sin{\beta} \\
d_{1} &= d_{\textbf{s}}\sin{\gamma} \\
d_{2} &= d_{\textbf{s}}\cos{\gamma} \\
d_{3} &= \sqrt{{r_{\textbf{s}}}^{2}-{d_{1}}^{2}} \\
d_{\textbf{h}} &= d_{2}-d_{3} \\
\end{aligned}
\label{input:figureld}
\end{equation}
Using this equation, we obtained $d_{\textbf{h}}$ which is the distance between $\textbf{O}_{camera}$ and highlight point position~$\textbf{h}$ that light is illuminated on the chrome ball using $d_{2}$ and $d_{3}$. Using $d_{\textbf{h}}$ and the coordinates of the highlighted point in the image plane, we obtained the coordinate~$\textbf{h}$ in the camera-based coordinate system. Additionally, we acquired $\textbf{R}$ and $\textbf{N}$ with coordinates by using Eq.~\ref{input:centerandhighlight}. In this equation, $\textbf{s}$ is the center of the chrome ball position and $\text{unit}$ is a function making vector to the unit normal vector.
\begin{equation}
\begin{aligned}
\textbf{R} &= (\textbf{h} - \textbf{O}_{camera})/d_{\textbf{h}} \\
\textbf{N} &= \text{unit}(\textbf{h} - \textbf{s})
\end{aligned}
\label{input:centerandhighlight}
\end{equation}
Finally, we obtained the light direction $\textbf{L}$ using Eq.~\ref{input:lightdirection}.
\begin{equation}
\textbf{L} = 2(\sum{\textbf{N} \odot \textbf{R}})\textbf{N} - \textbf{R}
\label{input:lightdirection}
\end{equation}

\begin{figure}[t]
    \centering
    \includegraphics[width=0.95\textwidth]{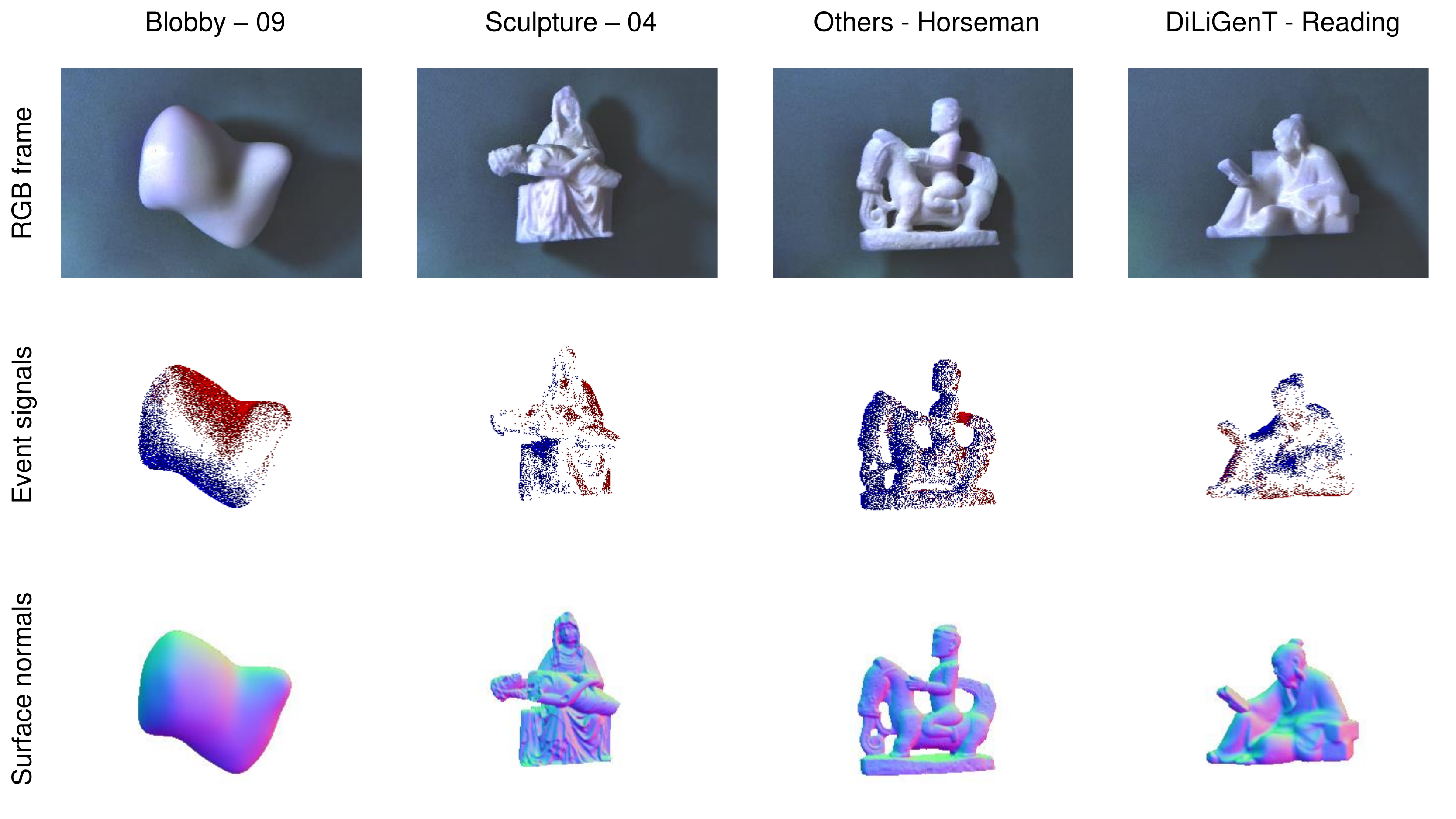}
    \caption{Samples of the datasets. This figure shows one of the 150 frames for each model, image event signals created with voxel grids, and ground truth surface normal map. All datasets are described in \ref{sec:datasets}.}
    \label{fig:dataset}
\end{figure}
\subsection{Train dataset}
Learning-based methods require training datasets to estimate the surface normal. We used 3D models introduced in \citep{chen2018ps} and \citep{ikehata2018cnn} as the training datasets. Among the 3D models introduced in CNN-PS, some 3D models that could not be accessed were replaced with other 3D models that had royalty-free licenses from the internet. Our training datasets are classified into three categories: smooth blobby, complex sculpture, and mixed others.

The Blobby dataset~\citep{johnson2011shape} is smooth overall, which implies that the depth difference of the surface changes smoothly over a wide range. Using only this dataset, it is difficult for the model to learn the detailed parts and exceptional intensity changes. To reduce bias, we used the Sculpture dataset~\citep{zisserman2017silnet} consisting of 307 sculptures with opposite properties. We employed a part of the 3D models that are introduced as the most complicated in \citep{chen2018ps}. Although we used an additional dataset for model learning, each dataset had obvious properties: smooth or complicated. For this reason, we included more 3D models with royalty-free licenses.

In conclusion, we gathered data under various conditions from three training datasets. Our training datasets help train the model to estimate the surface normal accurately, even with intensities impaired by ambient illumination. Some examples of training datasets given in Fig.~\ref{fig:dataset} are as follows; Blob09 in the Blobby dataset, Sculpture04 in the Sculpture dataset, and Horseman in the Others dataset. We applied a CLAHE image processing technique to enhance the clarity of the RGB frames and better show the shapes of the objects in the figure. The details of the datasets are provided in \ref{sec:datasets}. The training dataset excludes Blob05, Sculpture01, and Sculpture06, where large pixel errors occurred because of the mismatch in the translation from the RGB frames to the event camera plane.

\subsection{Test dataset}
The DiLiGenT dataset~\citep{shi2016benchmark} was created as a benchmark dataset for photometric stereo in the real world. The DiLiGenT dataset contains RGB frames, light information, and high-quality ground truth normals. These were acquired from 10 3D models with complex reflectance effects of various shapes to evaluate non-Lambertian photometric stereo. However, we require both RGB frames and event signal data. Our data acquisition system retrieves RGB frames, light information, high-quality ground truth normals, and even event signals using the 10 3D models in the DiLiGenT dataset. The DiLiGenT dataset was created in a darkroom environment, whereas the dataset we used was created in the wild environment where ambient light illumination exists. As part of the example, the reading 3D model is shown in Fig.~\ref{fig:dataset}.

\section{Experiments}
\subsection{Implementation details}
The output resolution of the event camera is $346 \times 260$ pixels, it is cropped to $208 \times 208$ pixels. However, the number of separate event voxel grids is one less than the number of frames because event signals are data generated while two frames are triggered. For this reason, we used the data made with 145 lights, excluding 5 lights from 150 lights, and the model constructed with PyTorch for the experiments. We trained EFPS-Net with Tesla-V100 GPU for 20 epochs with the adam optimizer, after setting the initial learning rate to 0.001, batch size to 2048, and resolution of the observation map to $32 \times 32$. Moreover, for EFPS-Net training, we used cosine annealing for the learning rate scheduler.

\subsection{Evaluation on test dataset}
\begin{figure}[t]
    \centering
    \includegraphics[width=0.95\textwidth]{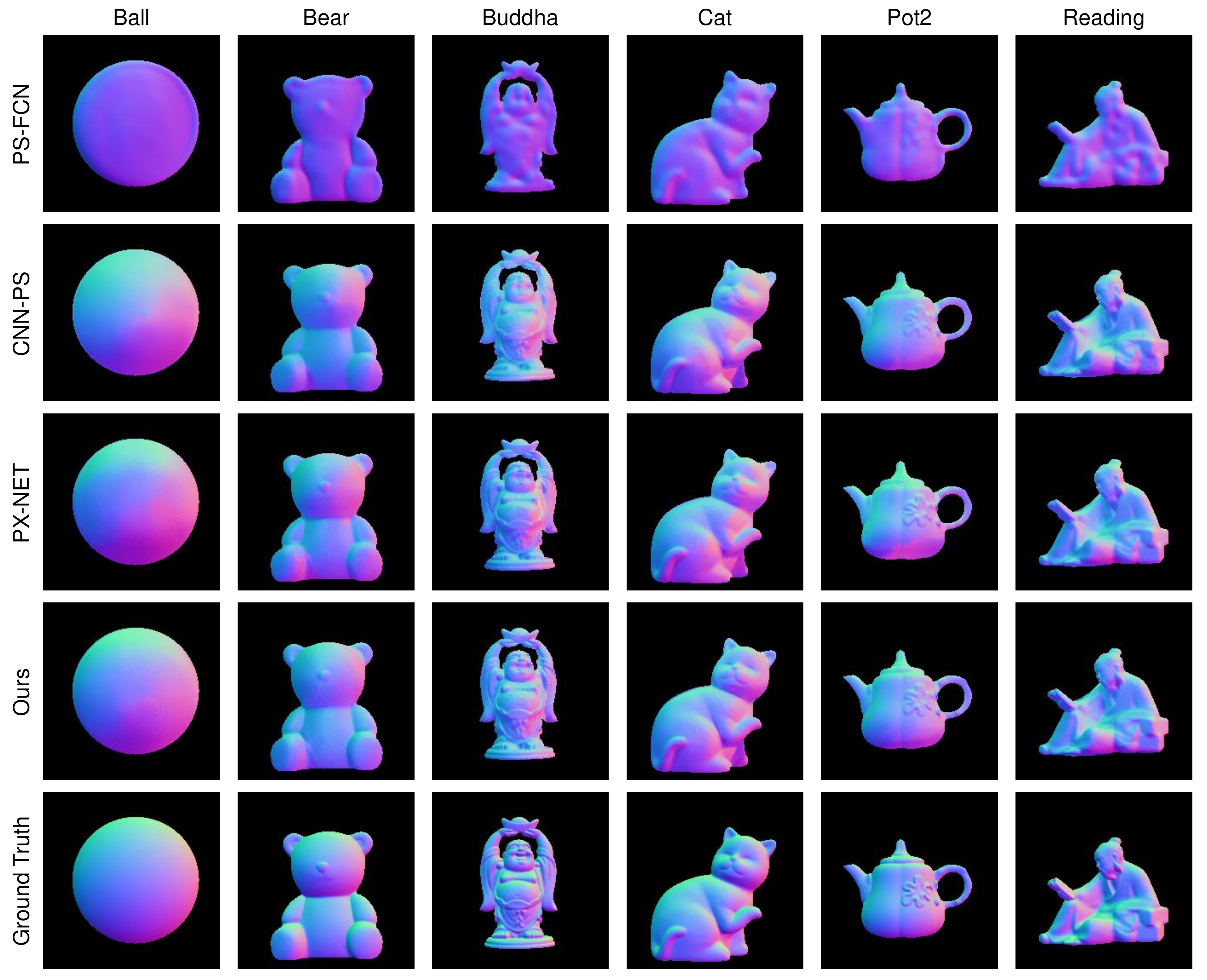}
    \caption{Qualitative result of learning-based methods on the RGB-event pair test dataset.}
    \label{fig:mainresult}
\end{figure}
\begin{table*}[!h]
\caption{Quantitative result of methods on the RGB-event pair test dataset. The underlined values are the second best-performing values, and the bolded values are the first best-performing values.}
\label{tab:all_result}
\resizebox{\textwidth}{!}{%
\renewcommand*{\arraystretch}{1.5}
\begin{tabular}{lccccccccccc}
\cline{1-12}
\multicolumn{1}{l}{Method} & Ball & Bear & Buddha & Cat  & Cow  & Goblet & Harvest & Pot1  & Pot2  & Reading & \multicolumn{1}{l}{Avg.}   \\

\hline
\multicolumn{1}{l}{N-LS~\citep{woodham1980photometric}} & 30.20 & 33.33 & 37.88 & 29.93 & 30.54 & 39.83 & 37.23 & 38.75 & 28.21 & 36.38 & \multicolumn{1}{l}{34.23} \\
\multicolumn{1}{l}{N-L1~\citep{ikehata2012robust}} & 30.31 & 31.88 & 37.18 & 29.63 & 29.33 & 39.24 & 36.49 & 37.13 & 27.98 & 36.55 & \multicolumn{1}{l}{33.57} \\
\multicolumn{1}{l}{N-SBL~\citep{ikehata2012robust}} & 30.99 & 31.61 & 37.57 & 29.83 & 30.23 & 40.37 & 38.03 & 37.97 & 28.26 & 36.77 & \multicolumn{1}{l}{34.06} \\
\multicolumn{1}{l}{PS-LS~\citep{ikehata2014photometric}} & 32.84 & 30.51 & 39.44 & 30.82 & 32.34 & 46.58 & 39.04 & 39.18 & 29.96 & 37.46 & \multicolumn{1}{l}{35.61} \\
\multicolumn{1}{l}{L-PLSBL~\citep{ikehata2014photometric}} & 36.25 & 31.40 & 39.72 & 31.80 & 31.20 & 45.53 & 38.41 & 38.23 & 31.81 & 37.51 & \multicolumn{1}{l}{36.19} \\
\cline{1-12}
\multicolumn{1}{l}{PS-FCN~\citep{chen2018ps}} & 34.54 & 36.26 & 38.05 & 35.01 & 30.64 & 30.23 & 31.51 & 30.38 & 29.19 & 30.39 & \multicolumn{1}{l}{32.62} \\
\multicolumn{1}{l}{CNN-PS~\citep{ikehata2018cnn}, K=10} & \ul{12.38} & 20.09 & 29.18 & 23.18 & 13.89 & 19.98 & \ul{29.12} & \ul{19.38} & \ul{15.56} & 22.84 & \multicolumn{1}{l}{20.56} \\
\multicolumn{1}{l}{PX-Net~\citep{logothetis2021px}, K=10} & 13.40 & \ul{17.96} & \ul{25.68} & \ul{21.11} & \ul{11.46} & \ul{17.61} & 29.56 & 19.84 & 15.94 & \ul{20.48} & \multicolumn{1}{l}{\ul{19.30}} \\
\cline{1-12}
\multicolumn{1}{l}{Ours, K=10} & \textbf{10.14} & \textbf{14.20} & \textbf{25.11} & \textbf{18.74} & \textbf{10.29} & \textbf{16.79} & \textbf{28.57} & \textbf{19.34} & \textbf{13.71} & \textbf{20.12} & \multicolumn{1}{l}{\textbf{17.71}} \\
\hline

\end{tabular}%
}
\end{table*}

Photometric stereo tasks conventionally use Mean Angular Error~(MAE) to estimate performance. Therefore, we used the average MAE as a metric, with a value indicated as an angular error. The angular error is the difference in the angle between the predicted normal and ground-truth normal corresponding to each pixel coordinate for an object. We performed experiments with RGB event pair datasets to prove that event signal data are effective for detecting even features. In particular, features are interrupted by ambient illumination in the real world. The MAE scores of the experimental results for each model are listed in Table~\ref{tab:all_result}. First, it is separated into three sections. The top section is the result of traditional methods, and the middle section is the result of learning-based methods. The methods in the two sections predict the surface normals using only RGB frames and light directions. By contrast, the bottom section describes our method of predicting surface normals using not only RGB frames and light directions but also event signals. Traditional methods achieved low prediction performance when changes were negligible. As shown in Fig.~\ref{fig:mainresult}, PS-FCN could not represent the surface normal maps when displayed in the tangent space. CNN-PS predicted surface normals appropriately, but the deviation in the normal values is small; therefore, CNN-PS cannot be represented in various ranges. Next, PX-Net predicted surface normals with more wide ranges, but PX-Net showed sharp changes when the color changed. Finally, EFPS-Net achieved similar ground-truth normals with spontaneous surface normal changes and state-of-the-art performance.

\subsection{Ablation studies}
This section discusses the experimental results of the two approaches, which contribute considerably to the performance improvement of EFPS-Net. The first is the rotational pseudo-invariant introduced in CNN-PS~\citep{ikehata2018cnn}, and the second is our model architecture. It is desirable for the MAE to be as close to 0 as possible.

\begin{figure}[t]
    \centering
    \includegraphics[width=0.95\textwidth]{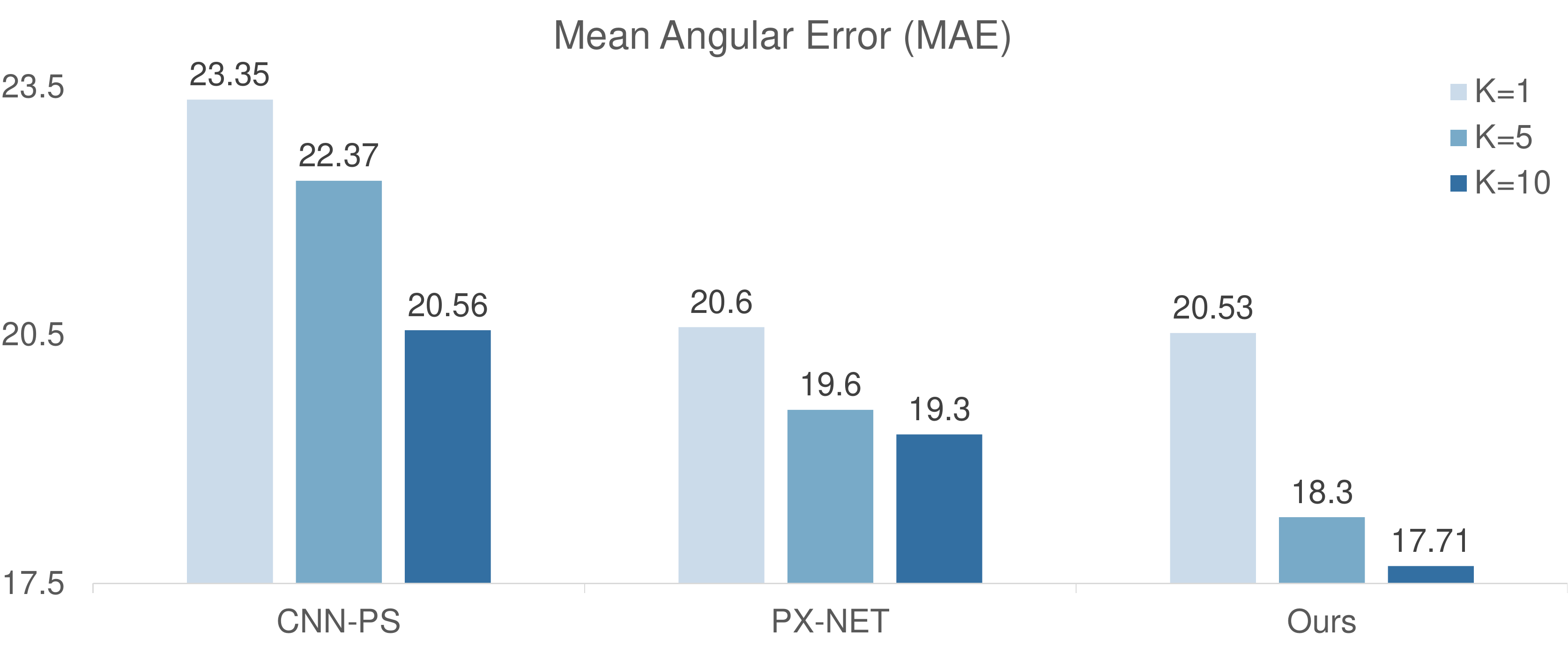}
    \caption{Mean Angular Error~(MAE) scores for each model with different rotational pseudo-invariances.}
    \label{fig:pseudoinvariance}
\end{figure}
\begin{table*}[!h]
\caption{To show the results on the effect of rotational pseudo-invariance. The bold values are the first best-performing values.}
\label{tab:ab1}
\resizebox{\textwidth}{!}{%
\renewcommand*{\arraystretch}{1.5}
\begin{tabular}{lccccccccccc}
\cline{1-12}
\multicolumn{1}{l}{Method} & Ball & Bear & Buddha & Cat  & Cow  & Goblet & Harvest & Pot1  & Pot2  & Reading & \multicolumn{1}{l}{Avg.}   \\

\hline
\multicolumn{1}{l}{CNN-PS~\citep{ikehata2018cnn}, K=1} & 18.68 & 24.58 & 29.26 & 29.75 & 15.29 & 22.74 & 31.18 & 20.85 & 17.55 & 23.62 & \multicolumn{1}{l}{23.35} \\
\multicolumn{1}{l}{CNN-PS~\citep{ikehata2018cnn}, K=5} & 14.27 & 21.60 & 30.79 & 24.31 & 15.95 & 24.44 & 30.74 & 21.88 & 15.60 & 24.07 & \multicolumn{1}{l}{22.37} \\
\multicolumn{1}{l}{CNN-PS~\citep{ikehata2018cnn}, K=10} & 12.38 & 20.09 & 29.18 & 23.18 & 13.89 & 19.98 & 29.12 & 19.38 & 15.56 & 22.84 & \multicolumn{1}{l}{20.56} \\
\cline{1-12}
\multicolumn{1}{l}{PX-Net~\citep{logothetis2021px}, K=1} & 13.23 & 17.53 & 27.48 & 22.08 & 11.38 & 21.79 & 30.02 & 22.01 & 19.76 & 20.66 & \multicolumn{1}{l}{20.60} \\
\multicolumn{1}{l}{PX-Net~\citep{logothetis2021px}, K=5} & 12.44 & 19.34 & 25.99 & 23.68 & 11.00 & 18.41 & 29.55 & 18.69 & 16.80 & 20.13 & \multicolumn{1}{l}{19.60} \\
\multicolumn{1}{l}{PX-Net~\citep{logothetis2021px}, K=10} & 13.40 & 17.96 & 25.68 & 21.11 & 11.46 & 17.61 & 29.56 & 19.84 & 15.94 & 20.48 & \multicolumn{1}{l}{19.30} \\
\cline{1-12}
\multicolumn{1}{l}{Ours, K=1} & 12.86 & 16.20 & 27.90 & 19.50 & 11.83 & 22.59 & 30.48 & 21.65 & 20.17 & 22.06 & \multicolumn{1}{l}{20.53} \\
\multicolumn{1}{l}{Ours, K=5} & 10.74 & \textbf{14.00} & 26.07 & \textbf{18.03} & 11.56 & 17.54 & 29.43 & 20.74 & 14.71 & 20.22 & \multicolumn{1}{l}{18.30} \\
\multicolumn{1}{l}{Ours, K=10} & \textbf{10.14} & 14.20 & \textbf{25.11} & 18.74 & \textbf{10.29} & \textbf{16.79} & \textbf{28.57} & \textbf{19.34} & \textbf{13.71} & \textbf{20.12} & \multicolumn{1}{l}{\textbf{17.71}} \\
\hline

\end{tabular}
}
\end{table*}
First, rotational pseudo-invariance augments data K times by rotating as much as $360\,^{\circ}$ divided by K, instead of interpolating the observation map. We set K to 1, 5, and 10 to confirm the extent to which this method assists. In Fig.~\ref{fig:pseudoinvariance}, all models indicate a performance improvement with as lower average MAE for all entire objects. As shown in Table~\ref{tab:ab1}, each model performs the best when K is 10. As a result, we verified that rotational pseudo-invariance leads to better performance by augmenting the data for a specific angle that could not be generated.

\begin{figure}[t]
    \centering
    \includegraphics[width=0.95\textwidth]{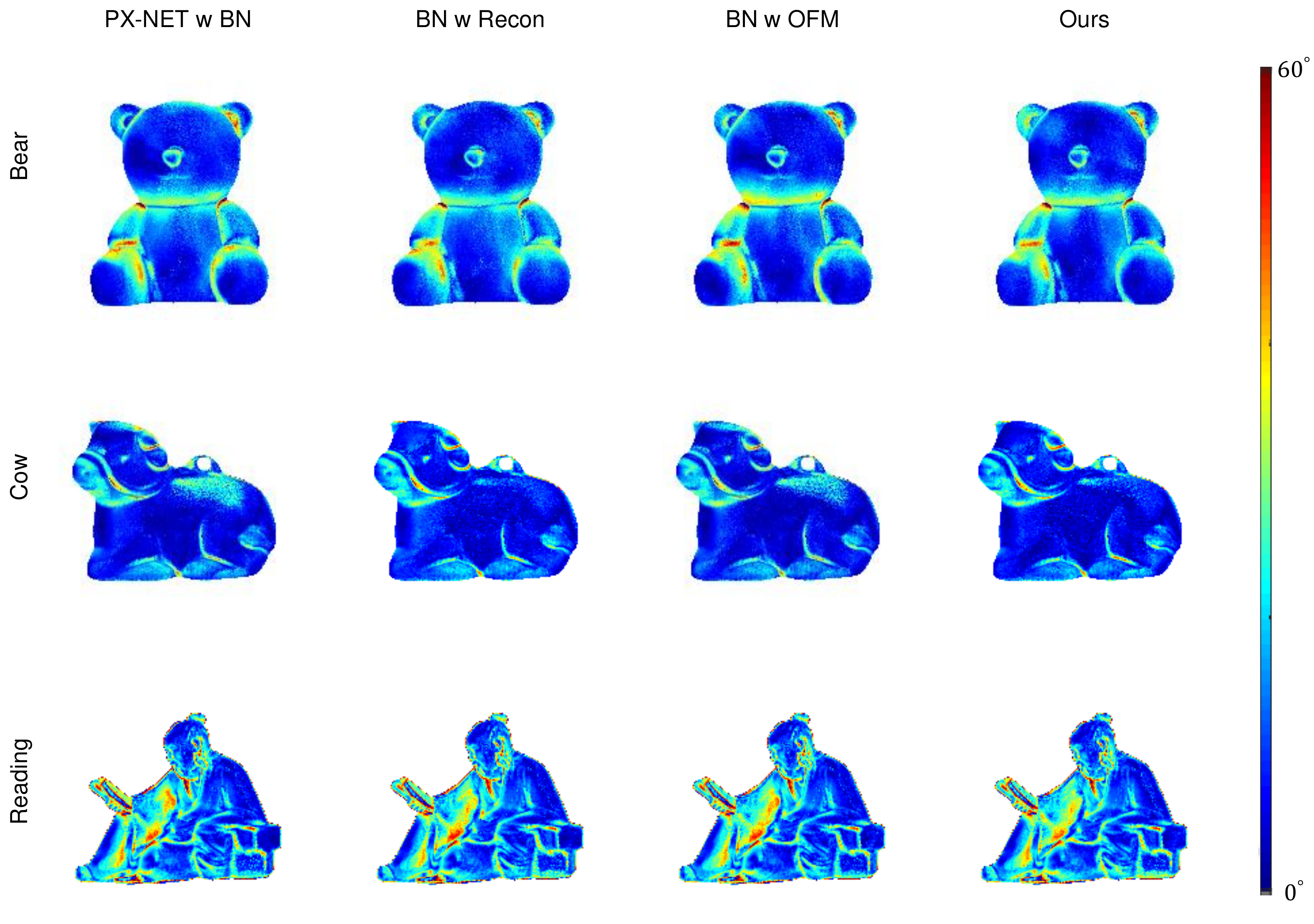}
    \caption{Visualization result of error maps for angular error to compare each of our methods. Each of our methods shows the result with three objects. The closer the error color is to red, the error value reaches an angle of $60\,^{\circ}$, and the closer it is to blue, the error value reaches an angle of $0\,^{\circ}$.}
    \label{fig:ab2result}
\end{figure}
\begin{table*}[!h]
\caption{Results of our methods on the RGB event pair test dataset. The bold values are the best values in terms of performance. The w/o $O_{e}$ indicates that $O_{e}$ are not contained as input, while w $O_{e}$ indicates that $O_{e}$ are included as input.}
\label{tab:ab2}
\resizebox{\textwidth}{!}{
\renewcommand*{\arraystretch}{1.5}
\begin{tabular}{lccccccccccc}
\cline{1-12}
\multicolumn{1}{l}{Method} & Ball & Bear & Buddha & Cat  & Cow  & Goblet & Harvest & Pot1  & Pot2  & Reading & \multicolumn{1}{l}{Avg.}   \\

\hline
\multicolumn{1}{l}{PX-Net~\citep{logothetis2021px} with BN~(w/o $O_{e}$), K=10} & 12.29 & 14.47 & 25.25 & 18.62 & 10.38 & 17.85 & 29.91 & 19.40 & 16.47 & 19.98 & \multicolumn{1}{l}{18.46} \\

\hline
\multicolumn{1}{l}{PX-Net~\citep{logothetis2021px} with BN~(w $O_{e}$), K=10} & 11.04 & 15.06 & 25.15 & 19.01 & 11.37 & 17.98 & \textbf{28.23} & 19.65 & 14.56 & 20.08 & \multicolumn{1}{l}{18.21} \\
\multicolumn{1}{l}{Ours with Recon, K=10} & 10.63 & 14.49 & 25.27 & \textbf{17.63} & 10.79 & \textbf{16.61} & 28.75 & 19.38 & 14.67 & 20.04 & \multicolumn{1}{l}{17.83} \\
\multicolumn{1}{l}{Ours with OFM, K=10} & 11.37 & 14.67 & 25.20 & 18.07 & 10.77 & 16.91 & 28.70 & \textbf{19.09} & 15.10 & \textbf{19.94} & \multicolumn{1}{l}{18.01} \\
\cline{1-12}
\multicolumn{1}{l}{Ours, K=10} & \textbf{10.14} & \textbf{14.20} & \textbf{25.11} & 18.74 & \textbf{10.29} & 16.79 & 28.57 & 19.34 & \textbf{13.71} & 20.12 & \multicolumn{1}{l}{\textbf{17.71}} \\
\hline

\end{tabular}
}
\end{table*}
Next, we conducted an experiment to determine the contribution of the components in our model architecture to improvements in model performance. To briefly introduce Table~\ref{tab:ab2}, PX-Net~\citep{logothetis2021px} with BN~(baseline model) represents Surface Normal Estimation Network~(SNE-Net) as the baseline model. Additionally, we provide qualitative results in Fig.~\ref{fig:ab2result}. This model uses observation maps concatenated with observation maps~$O_{r,g,b,n}$ generated from RGB frames and observation maps~$O_{e}$ generated from event signals as input data. Ours with Recon is an EFPS-Net approach, except Observation Fusion Module~(OFM). The last one, Ours with OFM, is an EFPS-Net approach, except for Event Interpolation Network~(EI-Net). Experimental results indicate that using a baseline model provides better performance, with lower MAE scores than that of PX-Net. Therefore, we confirm that using the batch normalization layer is beneficial in the case of a large batch size when acquiring feature maps from TB~(Transition Block). Moreover, Ours with recon showed greater performance, achieving a lower MAE score $0.38\,^{\circ}$ less than that achieved by using only the baseline model. However, in Ours with OFM, the MAE was $0.2\,^{\circ}$ lower than that of the baseline model. This result verifies that OFM fuses observation maps by compensating for each other. As shown in Fig.~\ref{fig:ab2result}, the errors are less distributed, especially in the angular part and boundary of the object. Finally, EFPS-Net achieved an MAE score $1.5\,^{\circ}$ less than that of the baseline model by appropriately using the advantages of the two components.

\section{Conclusion}
The objective of this study is to use the high dynamic range and low latency properties of event cameras in photometric stereo tasks to alleviate the darkroom setting problem and utilize requisite light information. Conventional RGB camera-based photometric stereo methods have limitations in adapting to an ambient light environment because of ambient illumination. Therefore, we propose EFPS-Net, which is the first approach to simultaneously accept RGB frames and event signals as input to account for the limitations of using only an RGB camera. Additionally, EFPS-Net trains with continuous light source trajectories, not discrete images, unlike the previous studies~\citep{santo2017deep,ikehata2018cnn,chen2018ps,logothetis2021px}. To prove that EFPS-Net addresses these constraints, we create a novel dataset including continuous light source trajectories. We demonstrate that our proposed method can estimate the surface normal of objects with decent quality under an ambient light environment and outperform other RGB camera-based state-of-the-art methods. 

\section*{Acknowledgements}
This research was fully supported by Culture, Sports, and Tourism R\&D Program through the Korea Creative Content Agency grant funded by the Ministry of Culture, Sports and Tourism in 2023 (Project Name: 4D Content Generation and Copyright Protection with Artificial Intelligence, Project Number: R2022020068)

\appendix

\section{Undistortion and Matching each plane}
\label{sec:equations}
\begin{equation}
\begin{aligned}
\textbf{I}&=\begin{bmatrix}
 f_{x} & \alpha & c_{x}  \\
 0 & f_{y} & c_{y}  \\
 0 & 0 & 1  \\
\end{bmatrix} \\
y_{n}&=(y_{1}-c_{y})/f_{y} \\
x_{n}&=(x_{1}-c_{x})/f_{x}-\alpha{y_{n}} \\
{r_{u}}^{2}&={x_{n}}^{2}+{y_{n}}^{2} \\
r_{d}&=1+k_{1}{r_{u}}^{2}+k_{2}{r_{u}}^{4}+k_{3}{r_{u}}^{6} \\
\begin{bmatrix}
 x_{un} \\
 y_{un} \\
\end{bmatrix}&=r_{d}\begin{bmatrix}
 x_{n} \\
 y_{n} \\
\end{bmatrix}+\begin{bmatrix}
 2p_{1}x_{n}y_{n}+p_{2}({r_{u}}^{2}+{2x_{n}}^{2}) \\
 2p_{2}x_{n}y_{n}+p_{1}({r_{u}}^{2}+{2y_{n}}^{2}) \\
\end{bmatrix} \\
\begin{bmatrix}
 x_{u} \\
 y_{u} \\
\end{bmatrix}&=\textbf{I}\begin{bmatrix}
 x_{un} \\
 y_{un} \\
\end{bmatrix}
\end{aligned}
\label{input:undistort}
\end{equation}
In Eq.~\ref{input:undistort}, the intrinsic matrix~$\textbf{I} \in \mathbb{R}^{3\times3}$ to capture in the image plane corresponding to the camera includes the focal length~$f_{x}, f_{y} \in \mathbb{R}$, principal point~$c_{x}, c_{y}  \in \mathbb{R}$, and skew coefficient $\alpha  \in \mathbb{R}$. Subsequently, we transformed the image plane of the RGB camera to the image plane of the event camera using the following equation:
\begin{equation}
\begin{aligned}
\textbf{P}_{e}&=\textbf{I}_{e}\textbf{E}_{e}=\begin{bmatrix}
 \textbf{P}^{1}_{e} & \textbf{P}^{2}_{e} & \textbf{P}^{3}_{e} & \textbf{P}^{4}_{e}  \\
\end{bmatrix} \\
\textbf{P}_{rgb}&=\textbf{I}_{rgb}\textbf{E}_{rgb}=\begin{bmatrix}
 \textbf{P}^{1}_{rgb} & \textbf{P}^{2}_{rgb} & \textbf{P}^{3}_{rgb} & \textbf{P}^{4}_{rgb}  \\
\end{bmatrix} \\
\begin{bmatrix}
 \hat{x}_{rgb} \\ \hat{y}_{rgb} \\ \psi  \\
\end{bmatrix}&=-\begin{bmatrix}
  &  & -x_{rgb}  \\
 \textbf{P}^{1}_{rgb} & \textbf{P}^{2}_{rgb} & -y_{rgb}  \\
  &  &  -1  \\
\end{bmatrix}^{-1} \textbf{P}^{4}_{rgb} \\
\begin{bmatrix}
 x_{e} \\ y_{e} \\ \Phi \\
\end{bmatrix}
&=\textbf{P}_{e}\begin{bmatrix}
 \hat{x}_{rgb} \\ \hat{y}_{rgb} \\ \ 0 \\ 1 \\
\end{bmatrix}
\end{aligned}
\label{input:planetransform}
\end{equation}
In this equation, $\psi$ and $\Phi$ are arbitrary unused values. To obtain the camera parameters to use the equations, we use \citep{zhang2000flexible} calibration algorithm using a checkerboard to obtain intrinsic matrix~$\textbf{I}$, extrinsic matrix~$\textbf{E}$, and non-linearity lens distortion coefficients~$k_{1}, k_{2}, k_{3}, p_{1}, \text{and}~p_{2}$ from each camera.

\section{Datasets}
\label{sec:datasets}
\centering\includegraphics[width=\textwidth]{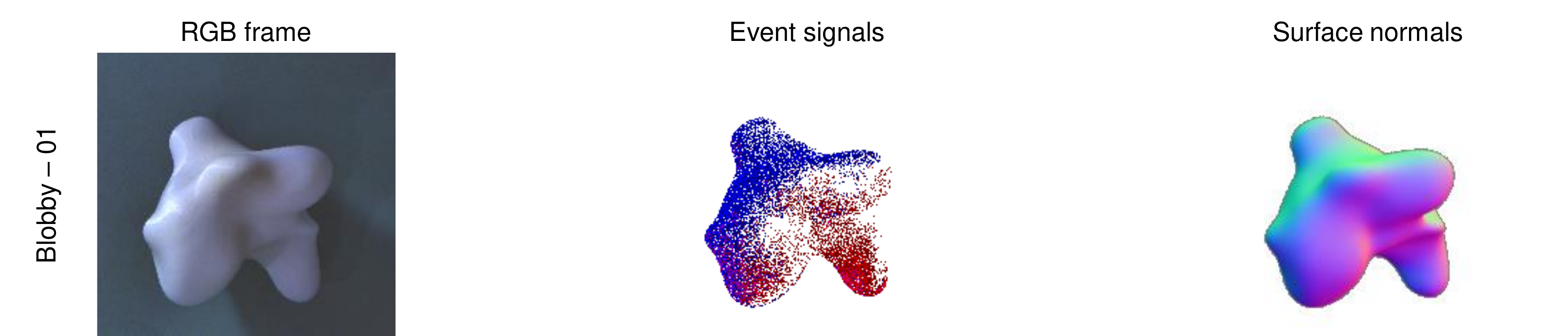}
\centering\includegraphics[width=\textwidth]{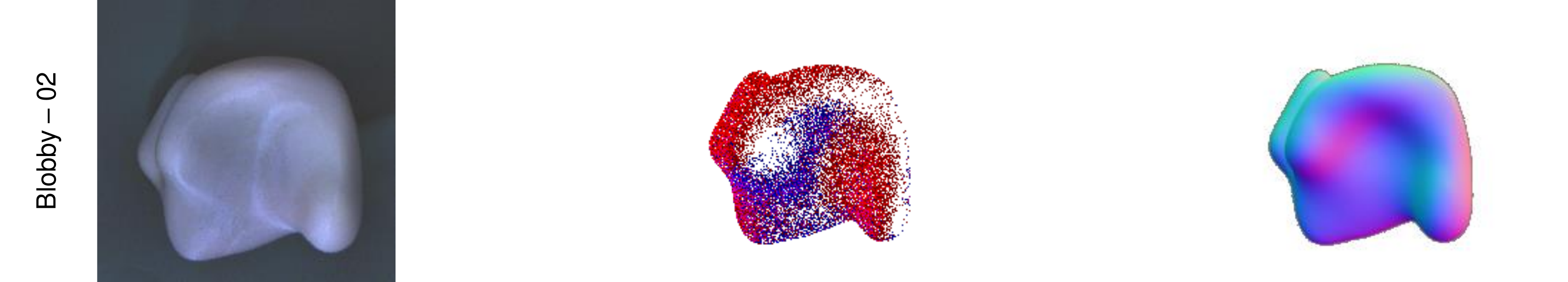}
\centering\includegraphics[width=\textwidth]{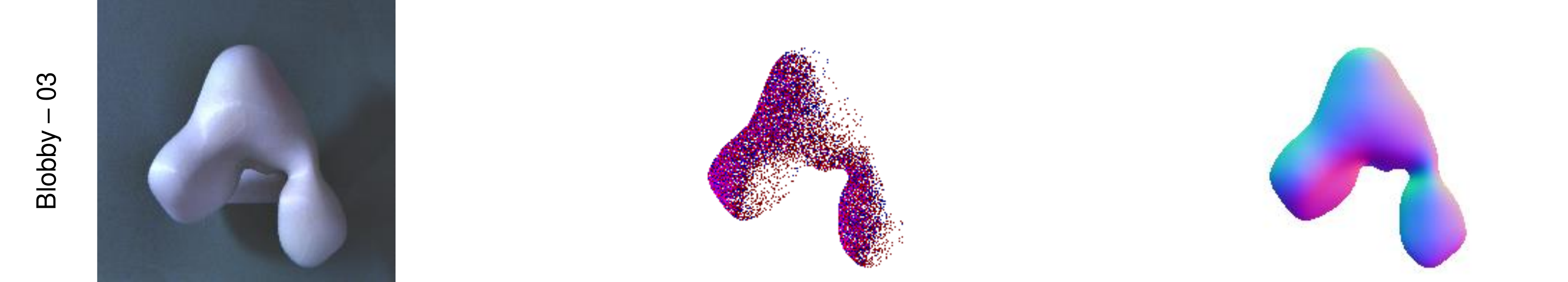}
\centering\includegraphics[width=\textwidth]{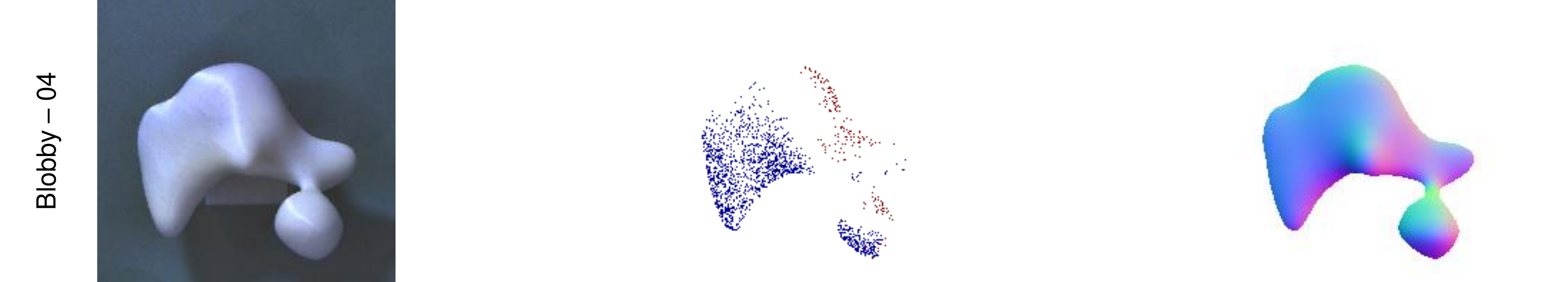}
\centering\includegraphics[width=\textwidth]{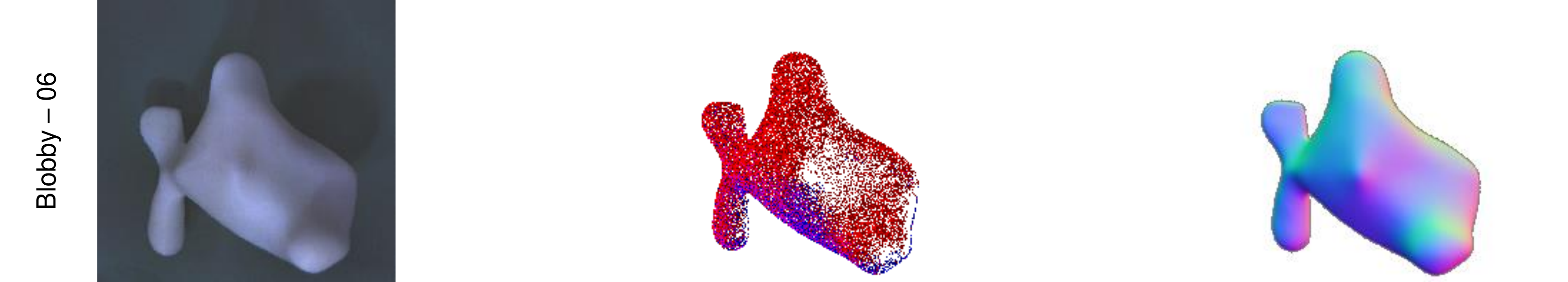}
\centering\includegraphics[width=\textwidth]{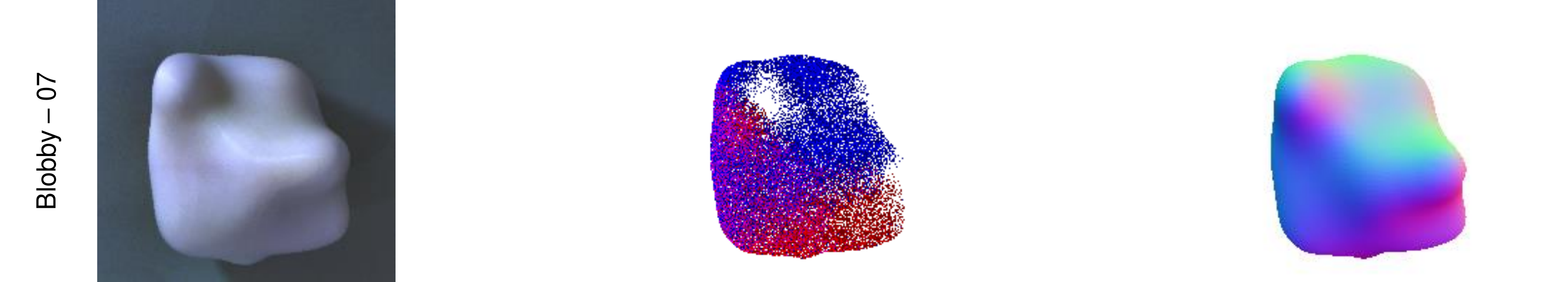}
\centering\includegraphics[width=\textwidth]{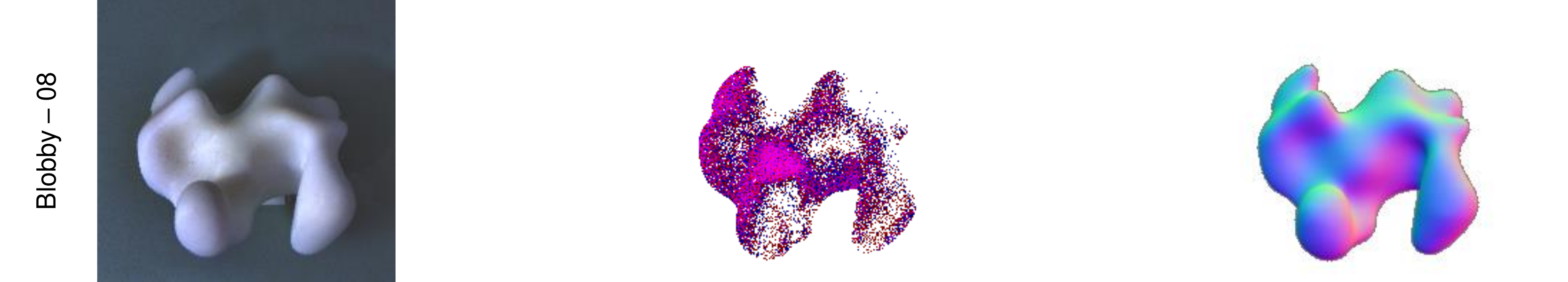}
\centering\includegraphics[width=\textwidth]{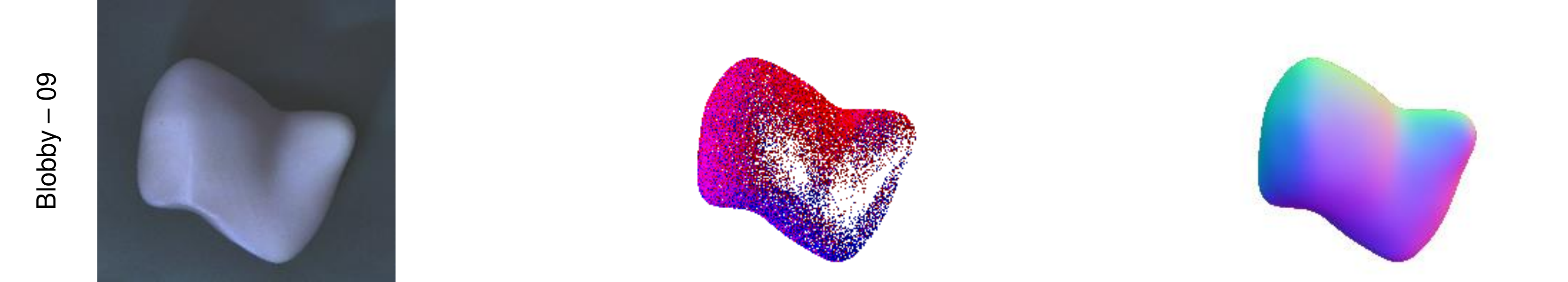}
\centering\includegraphics[width=\textwidth]{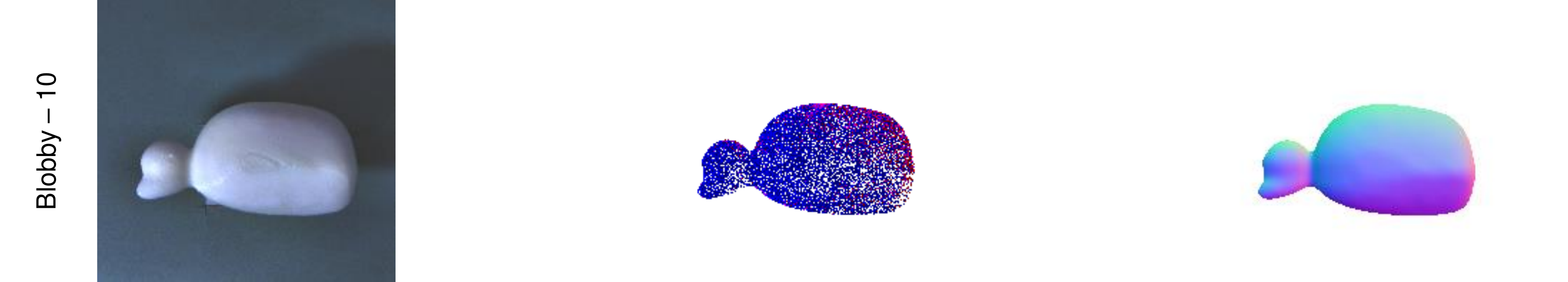}
\centering\includegraphics[width=\textwidth]{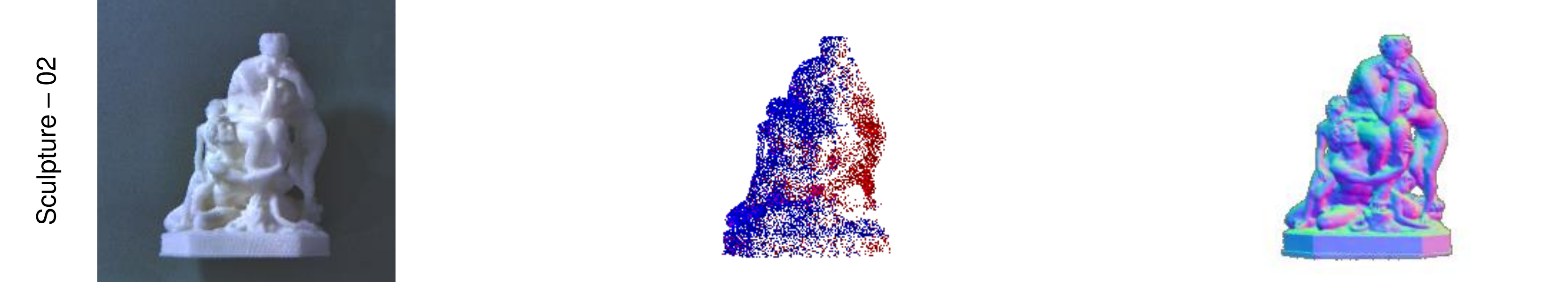}
\centering\includegraphics[width=\textwidth]{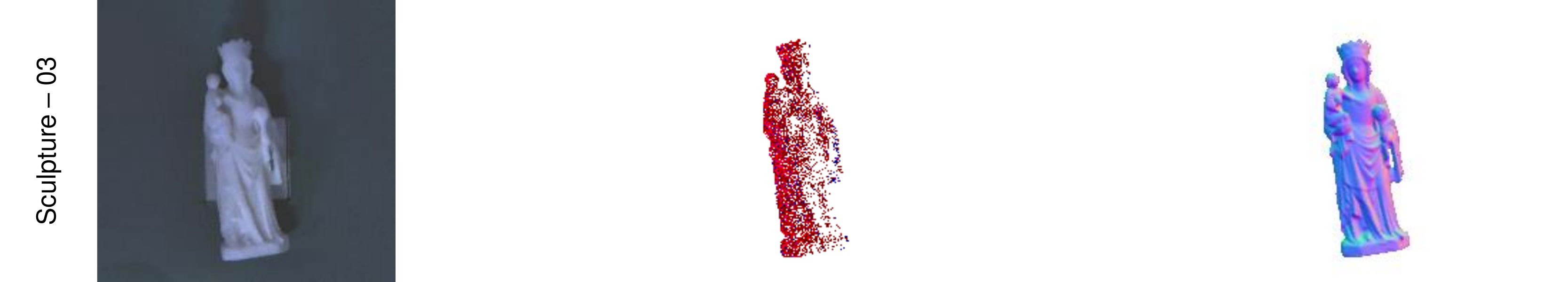}
\centering\includegraphics[width=\textwidth]{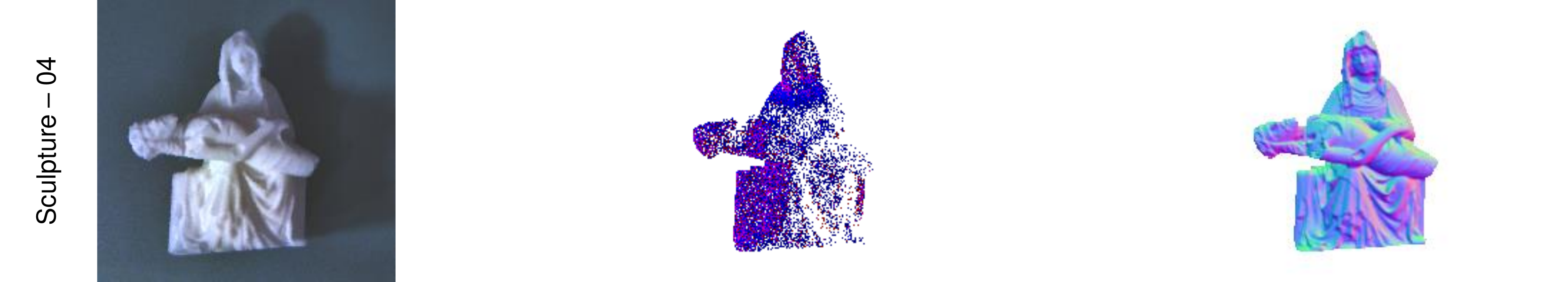}
\centering\includegraphics[width=\textwidth]{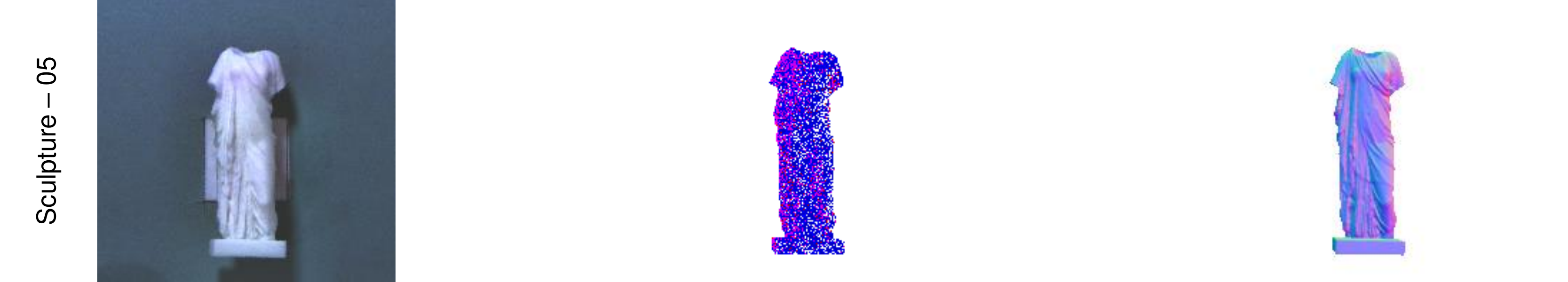}
\centering\includegraphics[width=\textwidth]{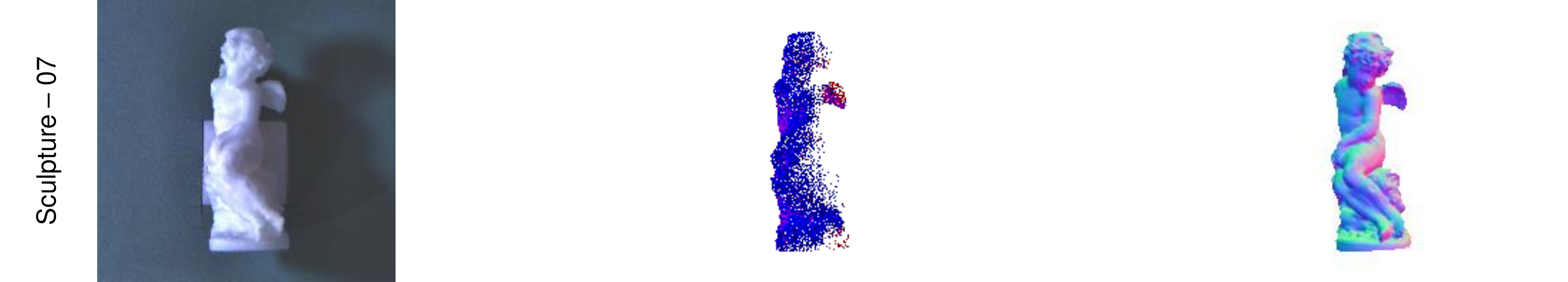}
\centering\includegraphics[width=\textwidth]{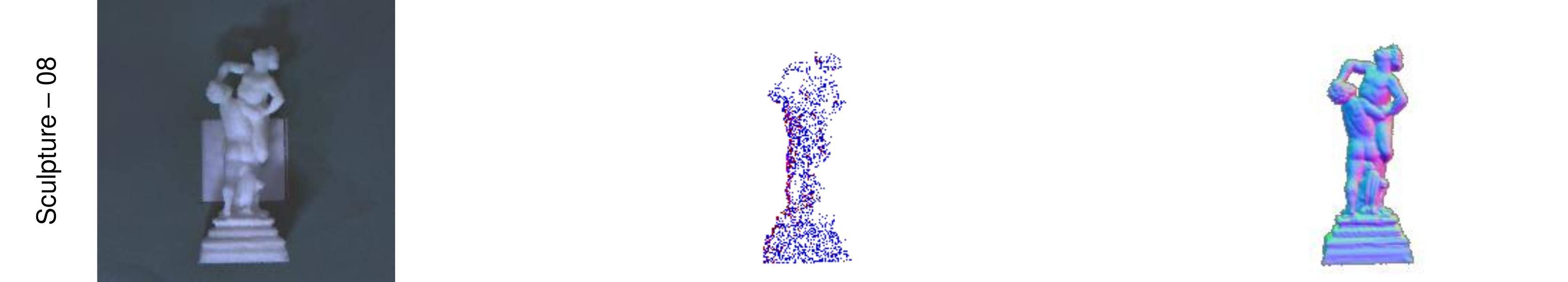}
\centering\includegraphics[width=\textwidth]{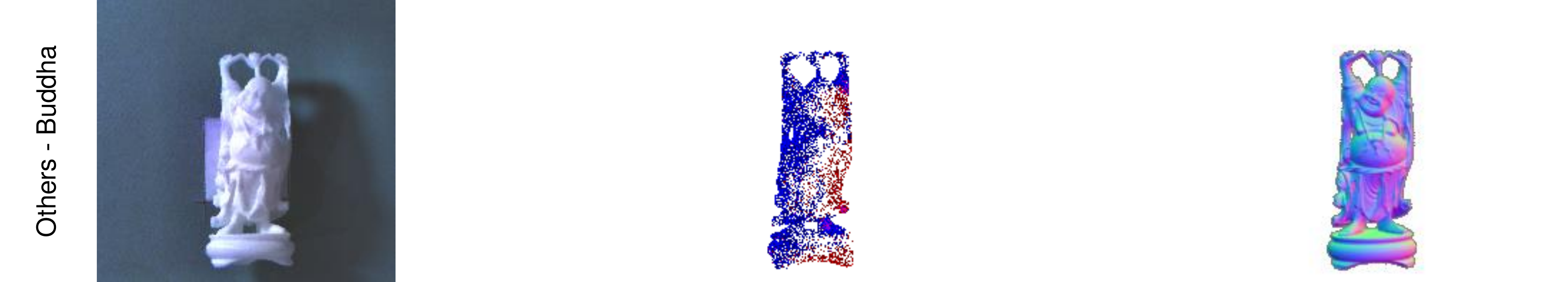}
\centering\includegraphics[width=\textwidth]{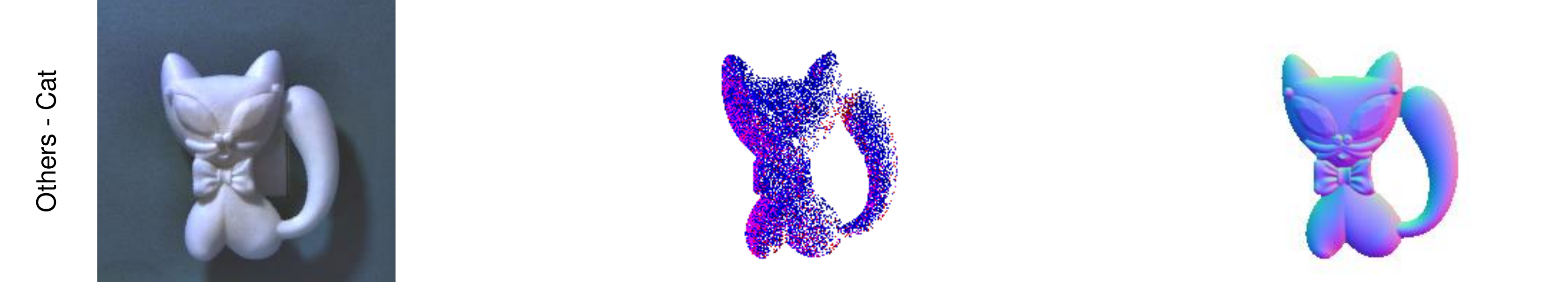}
\centering\includegraphics[width=\textwidth]{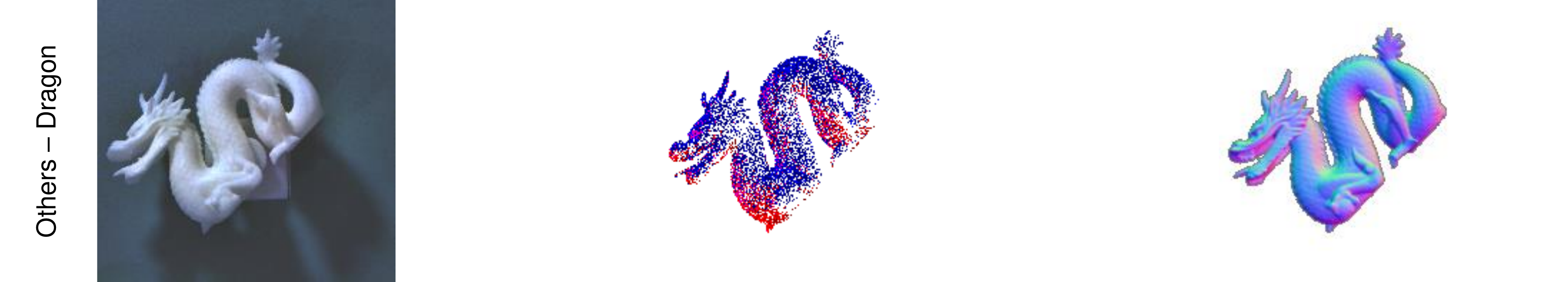}
\centering\includegraphics[width=\textwidth]{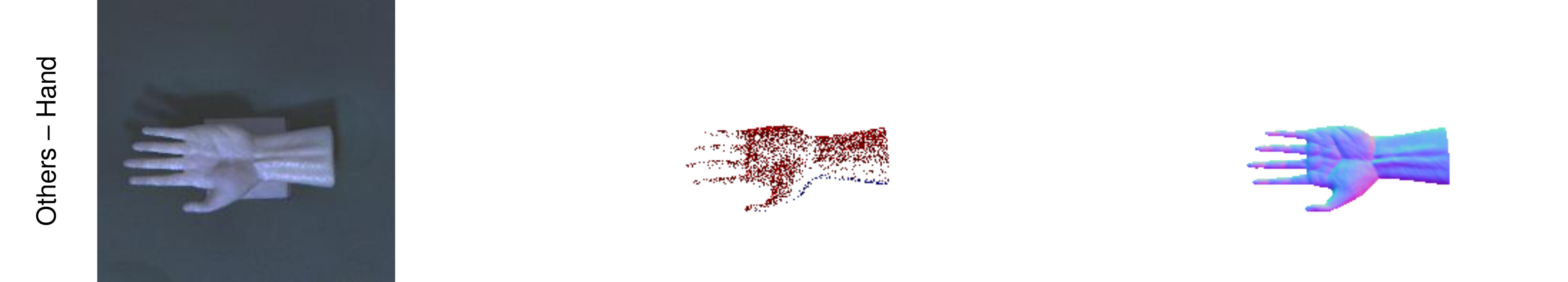}
\centering\includegraphics[width=\textwidth]{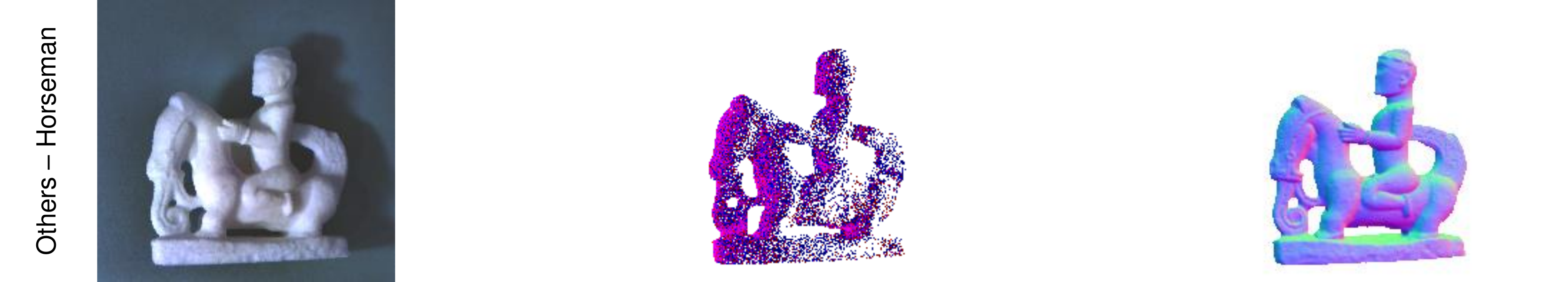}
\centering\includegraphics[width=\textwidth]{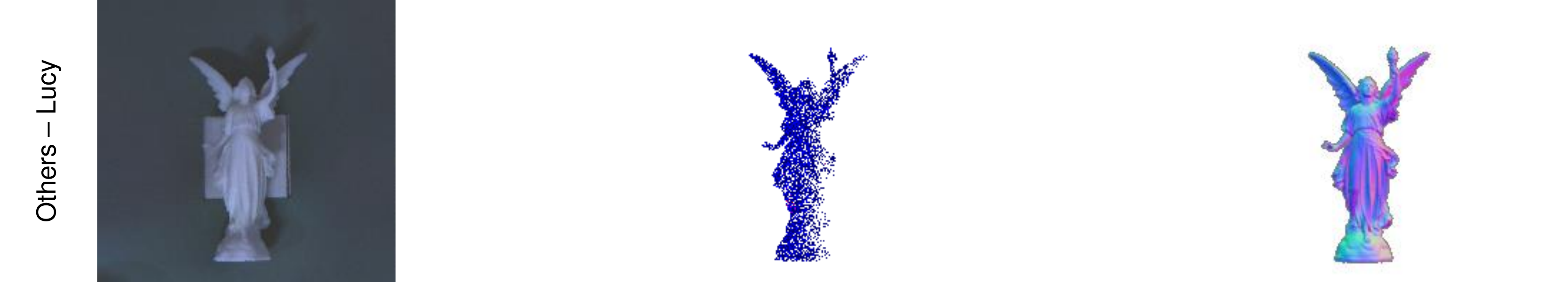}
\centering\includegraphics[width=\textwidth]{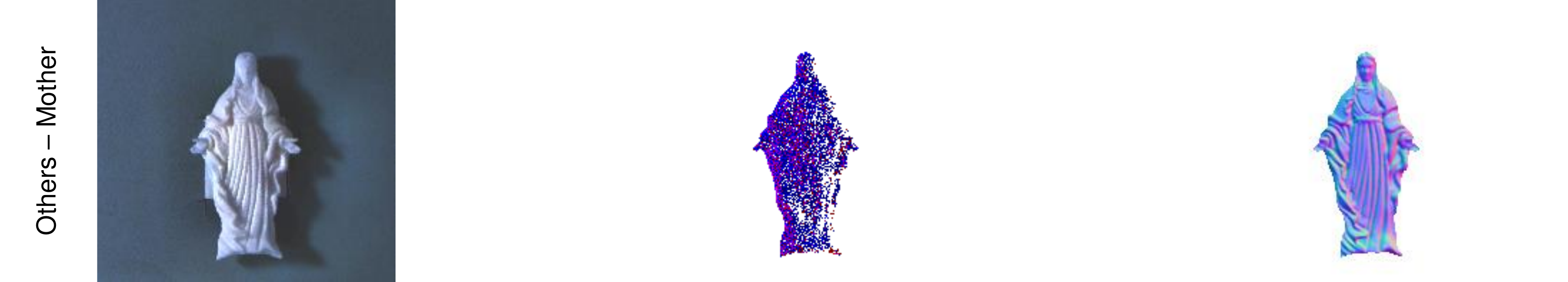}
\centering\includegraphics[width=\textwidth]{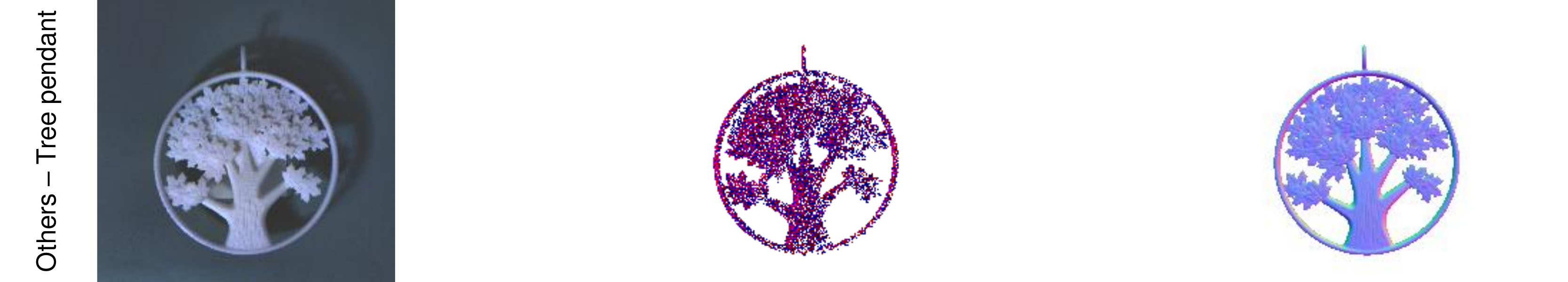}
\centering\includegraphics[width=\textwidth]{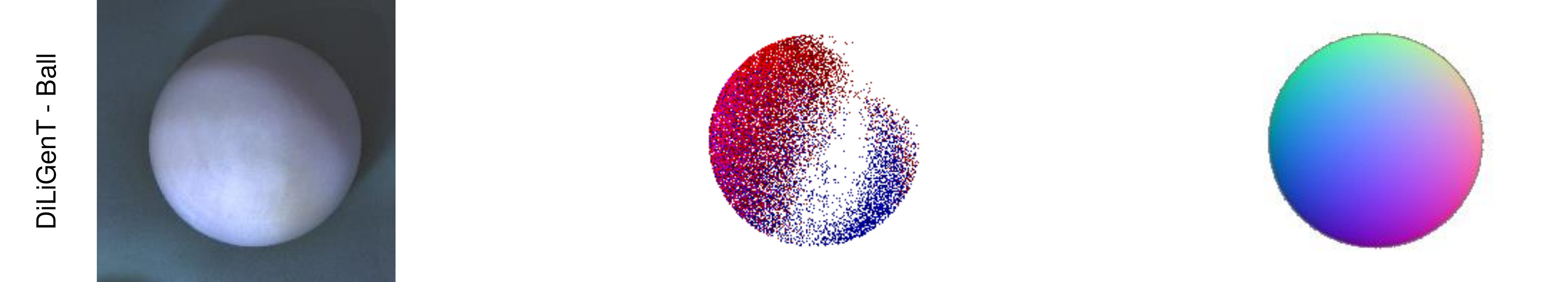}
\centering\includegraphics[width=\textwidth]{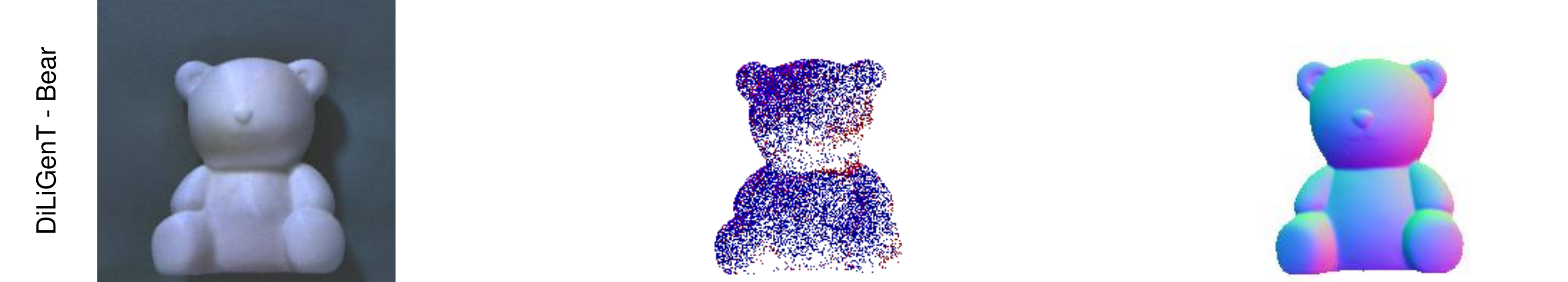}
\centering\includegraphics[width=\textwidth]{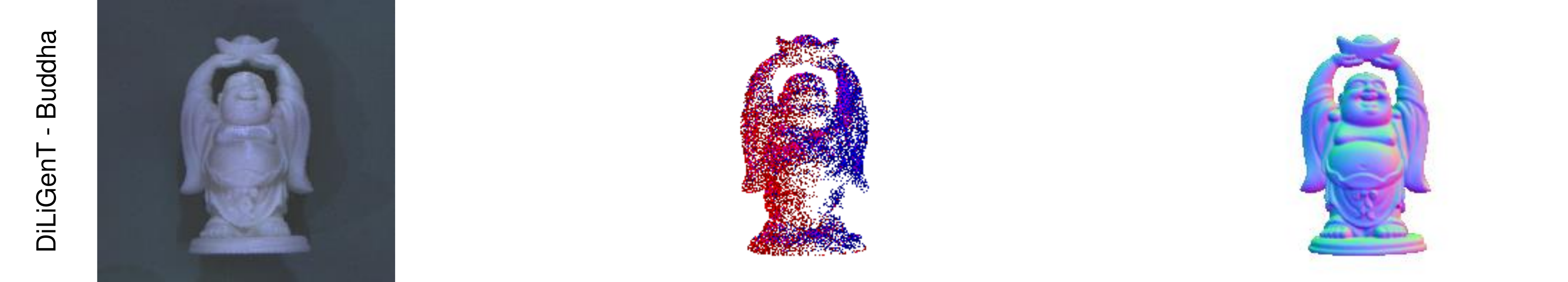}
\centering\includegraphics[width=\textwidth]{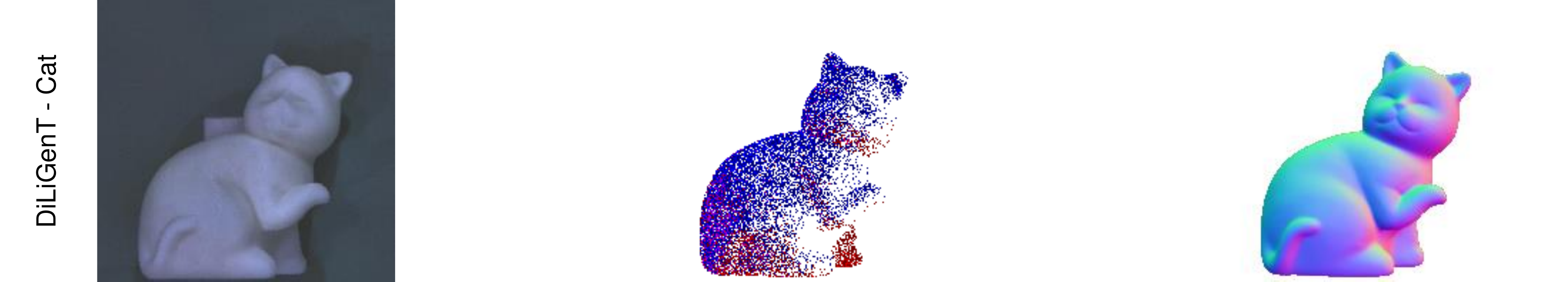}
\centering\includegraphics[width=\textwidth]{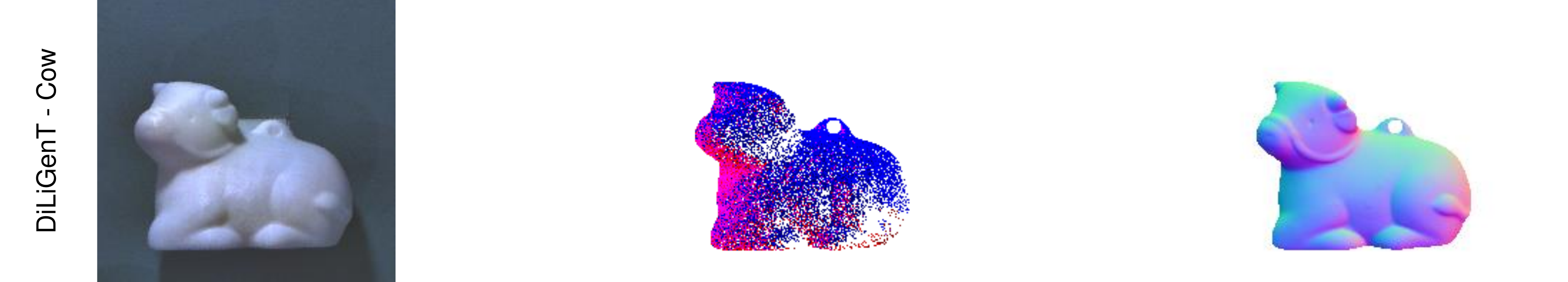}
\centering\includegraphics[width=\textwidth]{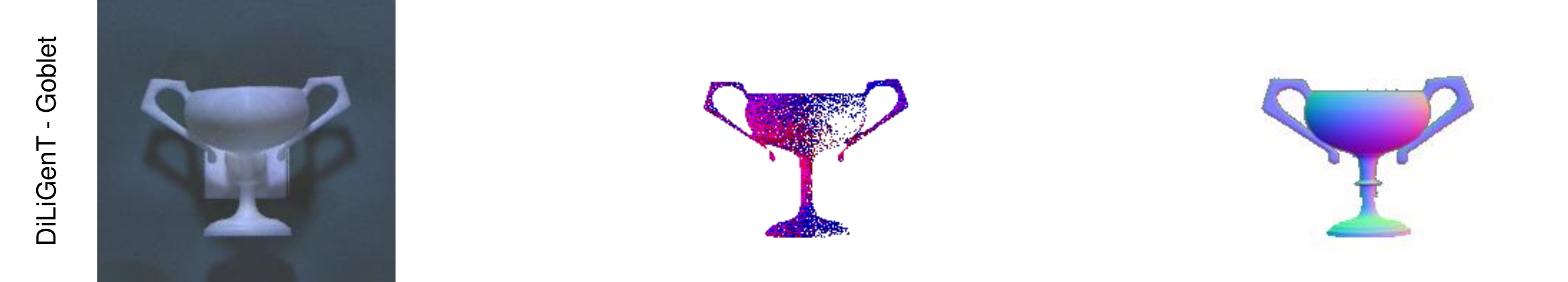}
\centering\includegraphics[width=\textwidth]{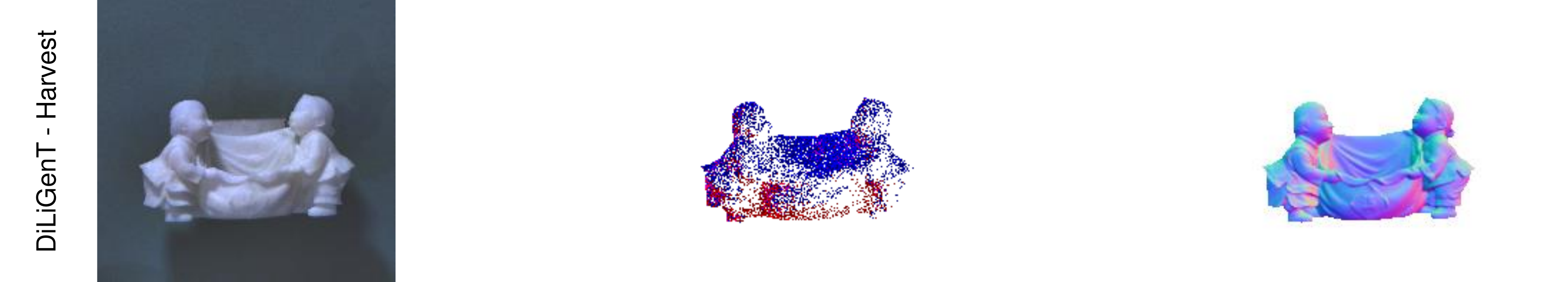}
\centering\includegraphics[width=\textwidth]{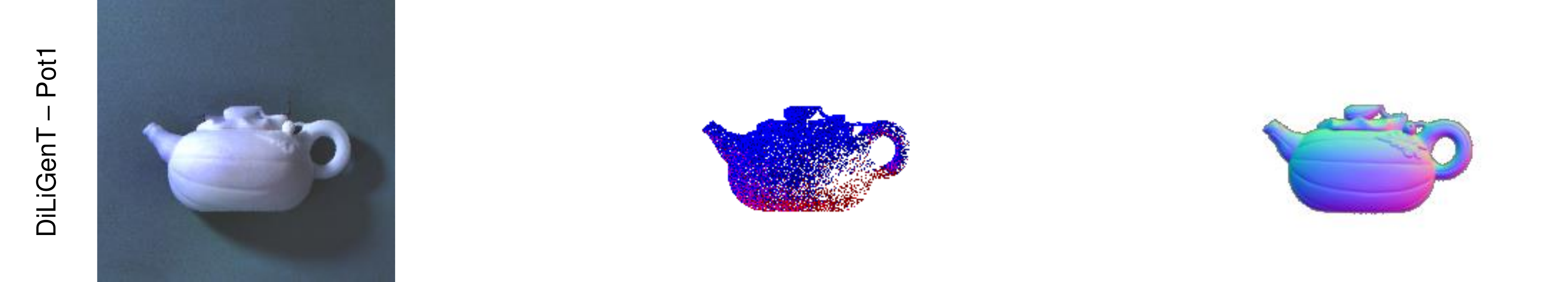}
\centering\includegraphics[width=\textwidth]{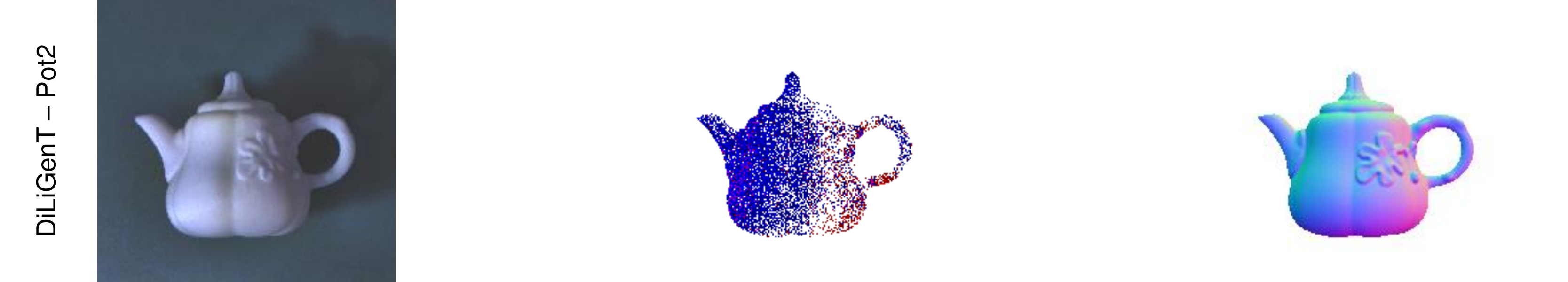}
\centering\includegraphics[width=\textwidth]{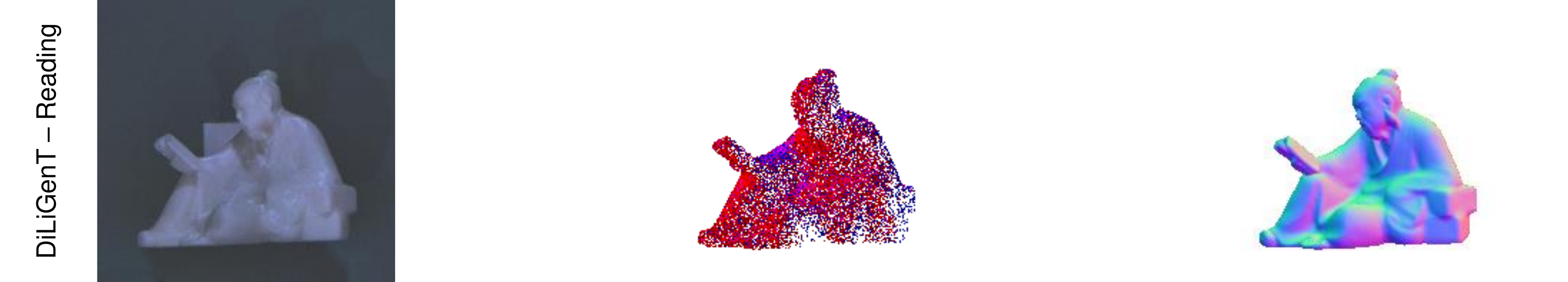}

\bibliographystyle{elsarticle-harv}
\bibliography{elsarticle}

\end{document}